%% file: main.tex
\documentclass[10pt,twocolumn,letterpaper]{article}

\usepackage{cvpr}              %

\input{preamble}
\definecolor{cvprblue}{rgb}{0.21,0.49,0.74}
\usepackage[pagebackref,breaklinks,colorlinks,allcolors=cvprblue]{hyperref}

\usepackage{times}
\usepackage{soul}
\usepackage{url}
\usepackage[utf8]{inputenc}
\usepackage{caption}
\usepackage{graphicx}
\usepackage{amsmath}
\usepackage{amsthm}
\usepackage{booktabs}
\usepackage{algorithm}
\usepackage{algorithmic}
\usepackage{amssymb}

\usepackage{subcaption}
\usepackage{multirow}

\usepackage{tcolorbox}

\def\paperID{37372} %
\def\confName{CVPR}
\def\confYear{2026}

\title{Mitigating Instance Entanglement in Instance-Dependent Partial Label Learning}

\author{Rui Zhao$^{12}$ \and Bin Shi$^{12}$\thanks{Corresponding author} \and Kai Sun$^{12}$ \and Bo Dong$^{23}$
\and 
$^{1}$School of Computer Science and Technology, Xi'an Jiaotong University\and
$^{2}$Shaanxi Province Key Laboratory of Big Data Knowledge Engineering, Xi'an Jiaotong University\and
$^{3}$School of Distance Education, Xi'an Jiaotong University
\and
{\tt\small rayn\_z@stu.xjtu.edu.cn, \{shibin,sunkai,dong.bo\}@xjtu.edu.cn}}

\begin{document}
\maketitle
\input{sec/0_abstract}
\input{sec/1_intro}
\input{sec/2_relatedwork}

\input{sec/3_method}
\input{sec/4_experiments}

\input{sec/5_conclusion}

\input{sec/appendix}

{
    \small
    \bibliographystyle{ieeenat_fullname}
    \bibliography{main}
}

\end{document}

%% file: sec/0_abstract.tex
\begin{abstract}
Partial label learning is a prominent weakly supervised classification task, where each training instance is ambiguously labeled with a set of candidate labels. In real-world scenarios, candidate labels are often influenced by instance features, leading to the emergence of instance-dependent PLL (ID-PLL), a setting that more accurately reflects this relationship. A significant challenge in ID-PLL is instance entanglement, where instances from similar classes share overlapping features and candidate labels, resulting in increased class confusion. To address this issue, we propose a novel Class-specific Augmentation based Disentanglement (CAD) framework, which tackles instance entanglement by both intra- and inter-class regulations. For intra-class regulation, CAD amplifies class-specific features to generate class-wise augmentations and aligns same-class augmentations across instances. For inter-class regulation, CAD introduces a weighted penalty loss function that applies stronger penalties to more ambiguous labels, encouraging larger inter-class distances. By jointly applying intra- and inter-class regulations, CAD improves the clarity of class boundaries and reduces class confusion caused by entanglement. Extensive experimental results demonstrate the effectiveness of CAD in mitigating the entanglement problem and enhancing ID-PLL performance. The code is available at https://github.com/RyanZhaoIc/CAD.git.
\end{abstract}

%% file: sec/1_intro.tex
\section{Introduction} \label{sec:intro}
Training reliable and effective Deep Neural Networks (DNNs) requires a large amount of accurately labeled data \cite{xia2022ambiguity}. However, due to data ambiguity and the high cost of annotation, obtaining precise labels remains challenging. To address this, many researchers have explored Partial Label Learning (PLL), a promising task for training DNNs using ambiguous labels, where each instance is associated with a set of candidate labels that includes the ground truth \cite{hullermeier2006learning}. Compared to precise labeling, these ambiguous labels can be collected more practically and cost-effectively through methods such as web mining \cite{xu2022webly}, crowdsourcing \cite{zhang2019multiple}, and automatic annotation \cite{briggs2012rank}. To learn from these candidate labels, label disambiguation, which involves identifying the true labels from these collected ambiguous labels, becomes the core challenge of PLL.

\begin{figure*}[t]
\centering
\includegraphics[width=0.9\linewidth]{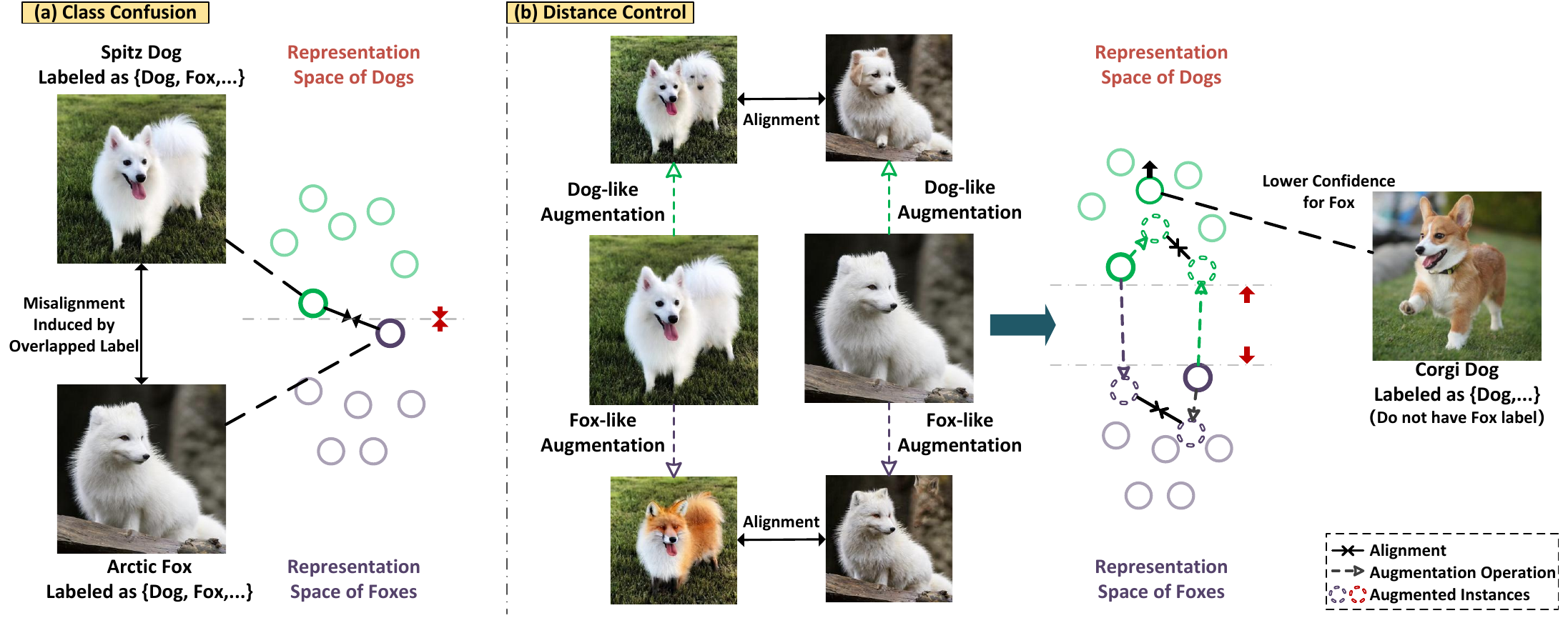} 
\caption{Illustration of how solely reducing intra-class distance may lead to class confusion (Figure (a)) and how class-specific augmentation combined with confidence-based penalties can mitigate confusion by increasing inter-class distance (Figure (b)). In Figure (a), similar instances with shared candidate labels may be misaligned. In Figure (b), class-specific augmentations are aligned while simultaneously lowering the confidence of confusing labels.}
\label{fig:intro_confusion}
\end{figure*}

Over the past decade, most works have assumed that the candidate labels are derived from \textit{instance-independent} noise, i.e., an incorrect label is assigned as a candidate label either randomly or with class-specific probabilities \cite{xia2022ambiguity,feng2020provably,wen2021leveraged,tian2024crosel,lyu2022deep,xie2021partial,wang2022pico}. However, in real-world scenarios, candidate labels are often related to the features of individual instances, which fits the assumptions of the \textit{Instance-Dependent Partial Label Learning} (ID-PLL) \cite{xu2021instance}. For example, even though both are dogs, annotators are more likely to mislabel a Spitz dog as a fox rather than mislabel a Corgi dog. 

To address ID-PLL, existing methods have introduced various robust learning theories and strategies, such as contrastive learning \cite{xia2022ambiguity}, Bayesian optimization \cite{xu2023progressive,qiao2022decompositional}, variational inference \cite{xu2024variational}, knowledge distillation \cite{wu2024distilling}, and causal graph inference \cite{li2024learning}. Among these, state-of-the-art methods \cite{xia2022ambiguity,wu2024distilling} leverage \textit{contrastive learning} \cite{he2020momentum} to enhance representations for inferring label information. Typically, in practice, they enhance representations by aligning instances predicted as the same class or sharing candidate labels, and then adjust label confidences using these improved representations for disambiguation, iteratively. This is grounded in the assumption that instances with similar representations are more likely to share the same ground-truth label \cite{wang2022pico}.

However, in ID-PLL, such intra-class alignment promoted by contrastive learning can be negatively affected by entangled instances, which share partially overlapping features and candidate labels. Emphasizing intra-class alignment without addressing this entanglement can exacerbate class confusion. For example, as shown in Figure \ref{fig:intro_confusion}(a), the Spitz dog and the Arctic fox are two closely entangled instances due to their visual similarity, with both being assigned the candidate labels ``fox'' and ``dog''. The similar features and overlapping labels make them prone to being considered as same-class instances, leading to misalignment. During training iterations, such misalignment continually reduces inter-class distance and provides incorrect disambiguation signals, ultimately increasing class confusion. Unfortunately, these cases are common in ID-PLL. For example, in widely used ID-PLL settings \cite{wu2024distilling} for Fashion-MNIST, CIFAR-10, and CIFAR-100, the most confused class pairs show that 97.67\%, 96.62\%, and 94.00\% of instances share their candidate labels, with 78.08\%, 75.09\%, and 7.80\% of instances having the same candidate labels.

To address this issue, we propose a Class-specific Augmentation based Disentanglement (CAD) framework, which tackles the instance entanglement problem through both intra- and inter-class regulation.
\textbf{For intra-class regulation}, we introduce a class-specific augmentation strategy to improve intra-class alignment. This strategy uses feature amplification techniques—such as feature attribution techniques or generative models—to enhance features related to each candidate label, generating augmentations closely aligned with the specific label. These augmentations guided by the same candidate label are then aligned using contrastive learning, preventing misalignment of features between similar classes. For example, as shown on the left side of Figure \ref{fig:intro_confusion}(b), a Spitz dog can be augmented for both the ``dog'' and ``fox'' classes by amplifying features related to each class. The dog-specific augmentations are then aligned with other instances' dog-specific augmentations, avoiding misalignment between traits of different classes.
\textbf{For inter-class regulation}, we introduce a weighted penalty loss that imposes stronger penalties on high-confidence negative classes. For example, as shown in Figure~\ref{fig:intro_confusion}(b), a Corgi dog may not carry the candidate label ``fox'' but can still exhibit high confidence in fox, due to their visual similarity. The proposed loss will penalize its confidence in fox with a higher weight, pushing the Corgi instance further away from the fox class and thereby reducing confusion between the two. In other words, the penalty helps amplify the discriminative signals of such typical instances—those with disjoint labels but similar features—which are particularly informative for distinguishing between similar classes. Overall, by handling the overlapping in both intra- and inter-class regulation, CAD can effectively increase inter-class distance and alleviate class confusion. The key contributions are as follows:
\begin{itemize}
    \item We focus on addressing class confusion caused by entangled instances, a problem with limited prior exploration in instance-dependent partial label learning.
    \item We propose a novel class-specific augmentation based disentanglement framework that mitigates the impact of entangled instances by intra- and inter-class regulation.
    \item Extensive experiments and analyses across various datasets demonstrate the effectiveness of our method.
\end{itemize}

%% file: sec/2_relatedwork.tex
\begin{figure*}[t]
    \centering
    \includegraphics[width=0.9\textwidth]{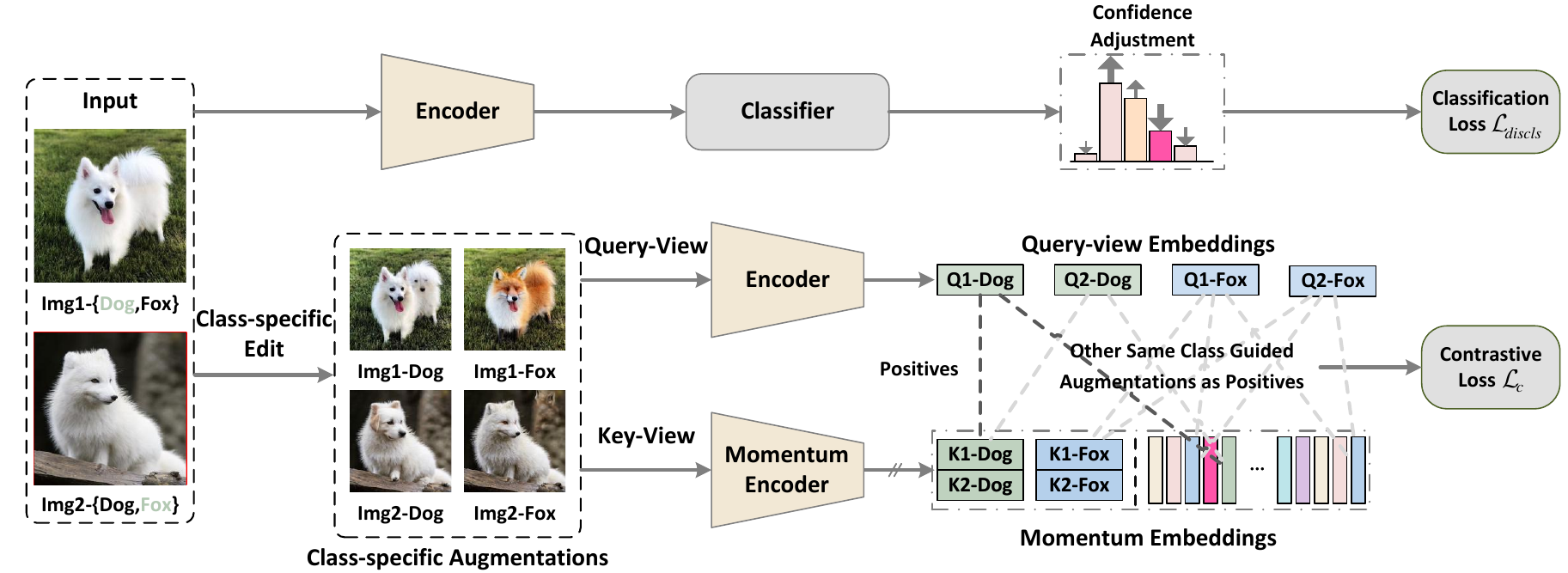} 
    \caption{Illustration of CAD framework. The upper part forces the model to produce outputs with less ambiguity based on weighted loss. The lower part generates class-specific augmentations that amplify specific features and guide them to have aligned representations. ``//'' denotes where gradient propagation is stopped. In the figure, ``Q1-Dog'' and ``K1-Dog'' represent the query and key embeddings for dog class-specific augmentations of img1, respectively. Here, augmentations guided by the same label are represented using the same color.}
    \label{fig:method}
\end{figure*}

\section{Related Work}
\label{sec:related}
PLL deals with ambiguous supervision in which each instance is associated with a candidate label set that includes the ground-truth label. Classical methods often treat labels independently of instance features, limiting their effectiveness in realistic settings. Recent works have therefore explored instance-dependent PLL \cite{qiao2022decompositional,xu2023progressive,he2023candidate}. For example, VALEN \cite{xu2021instance} infers posterior label distributions via a Dirichlet model; ABLE \cite{xia2022ambiguity} leverages instance-dependent cues through ambiguity-guided contrastive learning; and DIRK \cite{wu2024distilling} constructs reliable pseudo-labels by regulating the confidence of negative labels. However, these methods lack explicit treatment of similar-class confusion, which is pervasive in ID-PLL. Although early maximum-margin approaches \cite{yu2016maximum,nguyen2008classification} can implicitly enlarge distances of similar classes, they scale poorly to high-dimensional data and do not address entanglement from both representation and confidence perspectives. This section reviews recent advances in ID-PLL. For completeness, Appendix~\ref{asec:relateddiff} reviews additional literature on PLL, contrastive learning, diffusion-based image editing, and their relevance to our framework.

%% file: sec/3_method.tex
\section{Method}
In this section, we describe the proposed CAD framework in detail. The proposed method consists of two modules: representation learning for intra-class regulation (Section \ref{sec:rep}) and confidence adjustment for inter-class regulation (Section \ref{sec:cls}), as shown in Figure \ref{fig:method}.

\subsection{Problem Setting} \label{sec:problem_setting}
\subsubsection{Partial Label Learning}
Here, we define the problem of PLL in a formalized manner. Let $\mathcal{X} \subseteq \mathbb{R}^d$ be the $d$-dimensional feature space and $\mathcal{Y}=\{1,\dots,c\}$ the label space. A PLL dataset is given as $\mathcal{D}=\{(\boldsymbol{x}_i, \mathcal{S}_i)\}_{i=1}^n$, where $\boldsymbol{x}_i \in \mathcal{X}$ and $\mathcal{S}_i \subseteq \mathcal{Y}$ denote the $i$-th instance and its candidate label set, respectively, and $n$ denotes the number of samples. The latent true label $y_i$ of the instance $\boldsymbol{x}_i$ satisfies $y_i \in \mathcal{S}_i$. The task of PLL is to learn a classifier $f: \mathcal{X} \to \mathcal{Y}$ that maps each instance $\boldsymbol{x}_i$ to its corresponding ground truth $y_i$ from the training set $\mathcal{D}$, making label disambiguation within $\mathcal{S}_i$ the core challenge.

\subsubsection{Instance entanglement}
In this paper, we focus on mitigating the impact of entangled instances during training. Given a pair of candidate-labeled samples $(\boldsymbol{x}_i, \mathcal{S}_i)$ and $(\boldsymbol{x}_j, \mathcal{S}_j)$ with ground-truth labels $y_i$ and $y_j$, the instances $\boldsymbol{x}_i$ and $\boldsymbol{x}_j$ are considered entangled if the following conditions are jointly satisfied: (a) they belong to different classes; (b) each includes the ground-truth label of the other in its candidate label set; and (c) their feature representations are highly similar. Formally, the condition is defined as:
\begin{equation}
  (y_i \neq y_j) \land (\{y_i, y_j\} \subseteq \mathcal{S}_i \cap \mathcal{S}_j) \land \left(\operatorname{sim}(\boldsymbol{x}_i, \boldsymbol{x}_j) \geq \xi\right),
\end{equation}
where $\xi$ denotes a threshold that determines whether two instances are considered sufficiently similar under the similarity measure $\operatorname{sim}(\cdot, \cdot)$.

\begin{figure}[t]
    \centering
    \includegraphics[width=\linewidth]{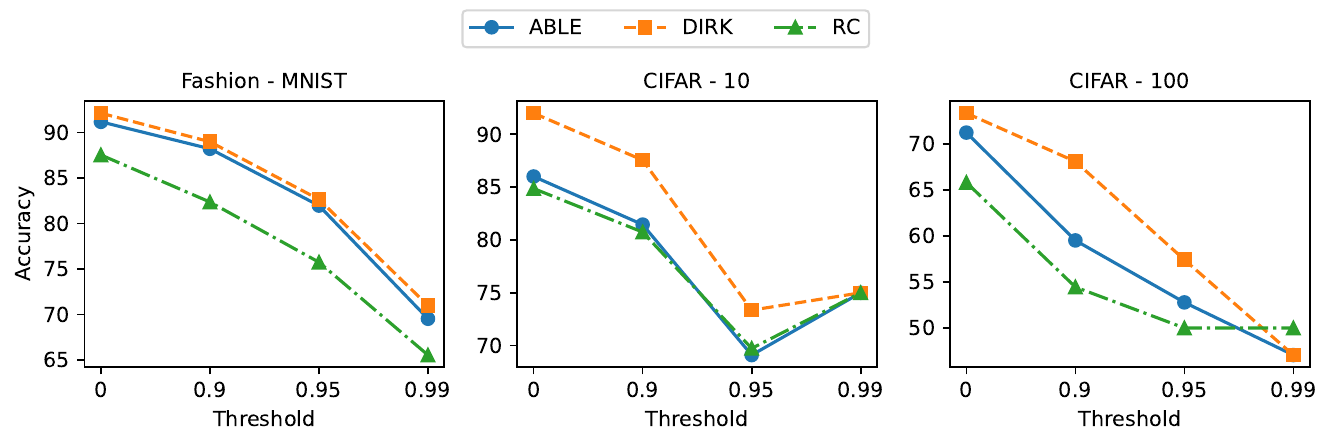} 
    \caption{Training accuracy on entangled instances under different similarity thresholds. ``0'' indicates that all entangled instances are used, reflecting the original accuracy. All models are trained on the full training set, and the entangled subsets are used for evaluation.}
    \label{fig:entan_xi_acc}
\end{figure}

To illustrate the prevalence and impact of entangled instances, we analyze three representative datasets—Fashion-MNIST \cite{xiao2017fashion}, CIFAR-10, and CIFAR-100 \cite{krizhevsky2009learning}. Table~\ref{tab:entan_xi_statistic} reports the statistics of these instances, and Figure~\ref{fig:entan_xi_acc} shows the training accuracy on them. We use a pretrained ResNet18 \cite{he2016identity} encoder to extract feature representations and cosine similarity to identify entangled pairs. As shown in Table~\ref{tab:entan_xi_statistic}, a substantial number of such pairs exist across datasets even at similarity thresholds above 0.90, indicating that they are both common and inherently difficult to distinguish. Figure~\ref{fig:entan_xi_acc} further shows that accuracy drops sharply as similarity increases, highlighting that existing methods struggle on these samples. To further verify that errors in Figure~\ref{fig:entan_xi_acc} stem from entanglement rather than intrinsic unlearnability, we report the fraction of misclassified entangled instances that are correct under full supervision in the Appendix \ref{asec:fraction}. These observations motivate our class-specific augmentation framework, which targets instance entanglement.

\begin{table}[t]
  \centering
  \caption{Statistics of entangled instances under different similarity thresholds. ``\#Pairs'' and ``\#Instances'' denote the number of entangled pairs and the number of unique instances involved, respectively. Note that one instance may appear in multiple pairs, so the counts are not in a one-to-one correspondence.}
  \label{tab:entan_xi_statistic}
  \resizebox{0.99\linewidth}{!}{
  \begin{tabular}{lllllllllllll}
    \toprule
      $\xi$ & \multicolumn{2}{c}{0.90} & \multicolumn{2}{c}{0.95} & \multicolumn{2}{c}{0.99} \\
      \cmidrule(lr){2-3}\cmidrule(lr){4-5}\cmidrule(lr){6-7}
      ~ & \#Pairs & \#Instances & \#Pairs & \#Instances & \#Pairs & \#Instances \\
      \midrule
      Fashion-MNIST & 528,996 & 38,717 & 125,535 & 16,992 & 3,658 & 1,802\\
      CIFAR-10 & 54,531 & 22,455 & 3,101 & 2,399 & 2 & 4 \\
      CIFAR-100 & 1,991 & 1,482 & 74 & 108 & 17 & 34 \\
    \bottomrule
  \end{tabular}
  }  
\end{table}

\subsection{Representation Learning for Intra-class Regulation}
\label{sec:rep}
To mitigate the susceptibility of representation distances to instance entanglement, we propose class-specific augmentations based on representation learning. Specifically, we first generate class-specific augmentations and then promote the alignment of augmentations that amplify features of the same class. This approach more effectively facilitates intra-class alignment by disentangling the features.

\subsubsection{Generating Class-specific Augmentations}
In ID-PLL, each candidate label corresponds to certain characteristics of an instance's features. Ideally, a class-specific augmentation should enhance the features associated with a specific class while preserving the instance's general characteristics. Given an instance $\boldsymbol{x}$ and its candidate label set $\mathcal{S}$, the class-specific augmentation $\boldsymbol{x}^\prime_s$ for class $s \in \mathcal{S}$ should satisfy the following constraints: (a) $\boldsymbol{x}^\prime_s$ should preserve the general characteristics of $\boldsymbol{x}$, formulated as $\min d(\boldsymbol{x}^\prime_s,\boldsymbol{x})$; (b) $\boldsymbol{x}^\prime_s$ should exhibit the representative features of class $s$, formulated as $\min d(\boldsymbol{x}^\prime_s,\bar{\boldsymbol{x}}^s)$. Here, $d(\cdot,\cdot)$ denotes a distance metric, and $\bar{\boldsymbol{x}}^s$ represents the most common feature of class $s$.

\smallskip\noindent\textbf{Instantiating the constraints with feature reweighting.} To operationalize these constraints, we first consider a lightweight instantiation based on Class Activation Mapping (CAM) \cite{zhou2016learning}, as it provides an efficient means to localize class-relevant features without requiring additional supervision. Specifically, for a given instance $\boldsymbol{x}$ and candidate label $s$, CAM can identify the features in $\boldsymbol{x}$ strongly activated by class $s$, formalized via a binary indicator vector $\boldsymbol{a}_s \in \{0,1\}^d$ (where $\boldsymbol{a}_{si}=1$ denotes activation for the $i$-th feature). To amplify class-$s$ features, we retain the activated components while scaling down non-activated ones via a blur parameter $\epsilon \in [0,1]$ as 
\begin{equation}
  \boldsymbol{x}^\prime_s = \boldsymbol{a}_s \odot \boldsymbol{x} + \epsilon \cdot (\boldsymbol{1} - \boldsymbol{a}_s) \odot \boldsymbol{x}.
\end{equation}
where $\odot$ denotes the Hadamard product. This emphasizes the salient features of class $s$ by attenuating non-activated features, thereby establishing our \textit{CAD-CAM} baseline. 

However, CAM-based augmentation is limited in its capacity to synthesize semantically coherent class details, as it merely reweights existing features without generating new ones. To address this limitation, we observe that the two core constraints—preserving instance identity and enhancing class-representative features—naturally align with the objective of conditional image editing: modifying an image to emphasize specific attributes while retaining its general characteristics.

\smallskip\noindent\textbf{Instantiating the constraints with feature editing.} Motivated by this insight, we further instantiate our framework using generative image editing, specifically through a novel application of InstructPix2Pix \cite{brooks2023instructpix2pix}, a diffusion-based model designed for instruction-guided image manipulation. While typically used for general-purpose edits, we repurpose it to produce class-specific augmentations tailored for ID-PLL. Concretely, we design a targeted augmentation strategy that uses the class name as an instruction to guide the editing process, ensuring that the generated sample highlights features associated with the corresponding label while preserving the instance's general structure. This process is formulated as $\boldsymbol{x}^\prime_s = E(\boldsymbol{x},I(s))$, where $E$ denotes the image editing model, and $I(s)$ is a function mapping class $s$ to its corresponding textual description. It should be noted that, although the cost of diffusion-based editing is higher than that of CAM-based augmentation, the process is performed offline and incurs only modest overhead ($\sim$24\% of total training time on CIFAR-10); see Appendix \ref{asec:training_cost} for details.

Together, the CAM-based and diffusion-based instantiations constitute a coherent spectrum of class-aware augmentations under the same core constraints. CAM provides an efficient, feature-level enhancement by selectively amplifying discriminative components, whereas diffusion-based editing enables richer, instance-specific transformations that synthesize semantically meaningful class-related details. Both approaches sharpen distinctions among candidate labels—enhancing intra-class alignment and mitigating inter-class confusion—thus highlighting the generality and versatility of the proposed CAD framework for PLL.

\subsubsection{Alignment of Class-specific Augmentations}
By generating class-specific augmentations, we obtain a set of augmented instances $\mathcal{A} = \bigcup_{(\boldsymbol{x}, \mathcal{S}) \in \mathcal{D}} \{\boldsymbol{x}'_s \mid s \in \mathcal{S}\}$, where $\boldsymbol{x}'_s$ denotes the augmentation of the instance $\boldsymbol{x}$ guided by its candidate label $s$. Within this set $\mathcal{A}$, augmentations induced by the same candidate label are treated as positive pairs, since they amplify the shared semantic features of that specific class (e.g., Img1-Dog and Img2-Dog in Figure \ref{fig:method}).

Following popular contrastive learning methods \cite{he2020momentum,khosla2020supervised,xia2022ambiguity}, we leverage these positive pairs for representation learning. Given an augmented sample $\boldsymbol{x}^\prime \in \mathcal{A}$ guided by label $y^\prime$, we first construct its query and key views $\alpha_q(\boldsymbol{x}^\prime)$ and $\alpha_k(\boldsymbol{x}^\prime)$ via distinct data transformations. These views are then encoded by query and momentum key encoders $e^q$ and $e^k$ into representations $\boldsymbol{q}_{\boldsymbol{x}^\prime}=e^q(\alpha_q(\boldsymbol{x}^\prime))$ and $\boldsymbol{k}_{\boldsymbol{x}^\prime}=e^k(\alpha_k(\boldsymbol{x}^\prime))$ (e.g., Q1-Dog and K1-Dog in Figure \ref{fig:method}). We then define the augmentation-based contrastive loss as
\begin{equation}
    \mathcal{L}_{c}(\boldsymbol{x}^\prime) = -\sum_{\boldsymbol{x}^+ \in \mathcal{A}_{y^\prime}}\mathrm{w}(\boldsymbol{x}^\prime,\boldsymbol{x}^+) \log{\mathrm{s}_{\tau}(\boldsymbol{q}_{\boldsymbol{x}^\prime},\boldsymbol{k}_{\boldsymbol{x}^+}, \mathcal{K})}, 
    \label{eq:lcon}
\end{equation}
where $\mathcal{A}_{y^\prime}$ denotes the set of augmentation samples induced by label $y^\prime$, $\mathrm{w}(\boldsymbol{x}^\prime,\boldsymbol{x}^+)$ denotes a weight that reflects the similarity between $\boldsymbol{x}^\prime$ and its aligned sample $\boldsymbol{x}^+$, and
\begin{equation}
  \mathrm{s}_{\tau}(\boldsymbol{q}_{\boldsymbol{x}^\prime},\boldsymbol{k}_{\boldsymbol{x}^+},\mathcal{K}) = \frac{\exp(\boldsymbol{q}_{\boldsymbol{x}^\prime}^\top \boldsymbol{k}_{\boldsymbol{x}^+}/\tau)}{\sum_{\tilde{\boldsymbol{k}} \in \mathcal{K}} \exp(\boldsymbol{q}_{\boldsymbol{x}^\prime}^\top \tilde{\boldsymbol{k}}/\tau)}
\end{equation}
denotes the temperature-scaled softmax over the key embedding set $\mathcal{K}=\{\boldsymbol{k}_{\boldsymbol{x}^\prime}|\boldsymbol{x}^\prime \in \mathcal{A}\}$ with temperature $\tau$. 

Here, the weight $\mathrm{w}(\boldsymbol{x}^\prime,\boldsymbol{x}^+)$ is introduced to account for the varying reliability of class-specific augmentations. Specifically, given the predicted logits $\boldsymbol{z}_{\boldsymbol{x}^\prime}$ and $\boldsymbol{z}_{\boldsymbol{x}^+}$ for positive pairs ($\boldsymbol{x}^\prime$, $\boldsymbol{x}^+$), the weight is then computed as a temperature-scaled softmax $\mathrm{w}(\boldsymbol{x}', \boldsymbol{x}^+) = \frac{\exp\left( \boldsymbol{z}_{\boldsymbol{x}'}^\top \boldsymbol{z}_{\boldsymbol{x}^+} / \tau_2 \right)}{\sum_{\tilde{\boldsymbol{z}} \in \mathcal{Z}_{y'}} \exp\left( \boldsymbol{z}_{\boldsymbol{x}'}^\top \tilde{\boldsymbol{z}} / \tau_2 \right)}$ over the positive sample logit set $\mathcal{Z}_{y^\prime}=\{\boldsymbol{z}_{\boldsymbol{x}^\prime}|\boldsymbol{x}^\prime \in \mathcal{A}_{y^\prime}\}$ with temperature $\tau_2$. This weighting mechanism upweights instance pairs whose augmented features more closely resemble the target class, while downweighting less reliable ones, thereby reducing the influence of noisy augmentations. In practice, we employ a confidence estimator $g^k: \mathcal{X}\rightarrow \mathbb{R}^c$, which shares the same backbone as the key encoder $e^k$, to produce the logits $\boldsymbol{z}_{\boldsymbol{x}^\prime}=g^k(\boldsymbol{x}^\prime)$. Both $g^k$ and $e^k$ are momentum-updated from their query-side counterparts—a classifier $g^q: \mathcal{X}\rightarrow \mathbb{R}^c$ and the query encoder $e^q$—where $g^q$ and $e^q$ share the same backbone, following the momentum update mechanism~\cite{he2020momentum,wu2024distilling}.

By aligning same-label-induced augmentations, we naturally address the main challenge in weakly-supervised contrastive representation learning: the reliable identification of positive pairs. Additionally, all class-specific augmentations derived from the same instance are included in representation learning. Because they originate from the same image but target different classes, they form semantically strong hard negatives that encourage the model to refine decision boundaries, especially for entangled classes with shared attributes.

\begin{table*}[ht]
    \centering
    \setlength{\tabcolsep}{12pt}
  \caption{Average accuracy and standard deviation (\%) over 5 runs. Best results are highlighted in \textbf{bold}, and the second best are \underline{underlined}.}
  \label{tab:accuracy}
    \resizebox{0.9\textwidth}{!}{
    \begin{tabular}{cccccc}
        \toprule
        & {Fashion-MNIST} & {CIFAR-10} & {CIFAR-100} & {Flower} & {Oxford-IIIT Pet}\\ 
        \midrule
        POP & 81.91 ± 0.37 & 89.55 ± 0.36 & 64.57 ± 0.37 & 33.02 ± 2.62 & 53.83 ± 1.55 \\
        VALEN & 82.91 ± 0.12 & 81.29 ± 0.39 & 60.19 ± 0.82 & 32.89 ± 0.88 & 50.74 ± 0.91 \\
        IDGP & 85.14 ± 0.39 & 84.12 ± 0.99 & 62.27 ± 1.89 & 41.26 ± 1.33 & 59.74 ± 0.71 \\
        CP-DPLL & 79.67 ± 6.95 & 80.29 ± 0.18 & 56.93 ± 1.47 & 42.42 ± 0.12 & 49.13 ± 2.49 \\
        CAVL & 78.06 ± 3.10 & 79.89 ± 0.07 & 59.76 ± 0.49 & 40.69 ± 1.93 & 50.64 ± 0.11 \\
        PICO & 85.92 ± 0.38 & 82.91 ± 1.43 & 58.56 ± 0.13 & 33.76 ± 1.30 & 62.64 ± 0.45 \\
        PRODEN &  83.77 ± 0.58 & 83.18 ± 0.29 & 67.77 ± 0.38 & 41.28 ± 0.57 & 52.77 ± 0.66 \\
        LWS & 75.52 ± 0.29 & 78.32 ± 2.11 & 65.74 ± 0.60 & 21.24 ± 0.63 & 31.21 ± 0.47 \\
        RC & 84.87 ± 1.48 & 87.53 ± 0.94 & 65.26 ± 0.40 & 41.20 ± 0.23 & 54.01 ± 0.05 \\
        CC & 79.98 ± 0.01 & 82.14 ± 0.01 & 62.28 ± 0.01 & 40.92 ± 0.40 & 52.99 ± 1.20 \\ 
        CEL & 87.78 ± 0.85 & 89.18 ± 0.40 & 68.73 ± 0.16 & 38.51 ± 1.02 & 68.19 ± 1.62 \\
        ABLE & 89.81 ± 0.08 & 83.92 ± 0.67 & 63.92 ± 0.39 & 43.51 ± 0.93 & 54.19 ± 1.01 \\
        DIRK & 91.48 ± 0.21 & 90.87 ± 0.25 & 68.77 ± 0.49 & 44.03 ± 0.02 & 64.95 ± 2.11 \\
        \midrule
        CAD-CAM & \underline{91.64 ± 0.29} & \underline{92.69 ± 0.10} & \underline{69.08 ± 1.84} & \textbf{49.67 ± 2.84} & \textbf{74.56 ± 3.16}  \\
        CAD & \textbf{92.14 ± 0.91} & \textbf{93.57 ± 0.24} & \textbf{72.03 ± 1.88} & \underline{47.88 ± 1.89} & \underline{69.46 ± 2.35}  \\ 
        \bottomrule
\end{tabular}}
\end{table*}

\subsection{Confidence Adjustment for Inter-class Regulation}
\label{sec:cls}
Through representation learning combined with class-specific augmentations, we mitigate the intra-class misalignment problem. To further increase inter-class distance, we employ an extended form of the weighting loss \cite{feng2020provably,wen2021leveraged}, which incorporates penalties for non-candidate classes that are highly confusable with the candidate classes.

Given a candidate-labeled sample $(\boldsymbol{x}, \mathcal{S})$, the set of non-candidate labels is defined as $\bar{\mathcal{S}} = \mathcal{Y} \setminus \mathcal{S}$. Our objective is to amplify high-confidence supervision signals in $\mathcal{S}$ while suppressing high confidence signals in $\bar{\mathcal{S}}$. Representing $\mathcal{S}$ as a $c$-dimensional indicator vector $\boldsymbol{s}$, where the $i$-th component $\boldsymbol{s}_i = 1$ indicates that label $i \in \mathcal{S}$ is present (and $0$ otherwise), the disambiguation classification loss is
\begin{equation}
    \mathcal{L}_{discls}(\boldsymbol{x})=\sum_{j \in \mathcal{Y}}\omega_j\ell(\boldsymbol{s}_j, \boldsymbol{x}),
    \label{eq:dc}
\end{equation}
where $\ell(\boldsymbol{s}_j,\boldsymbol{x})
= \boldsymbol{s}_j\,\psi(g^q_j(\boldsymbol{x}))
+ (1-\boldsymbol{s}_j)\,\psi(-g^q_j(\boldsymbol{x}))$, with $\psi(\cdot)$ is a non-increasing binary loss (e.g., the sigmoid loss), and $g^q_j(\boldsymbol{x})$ is the $j$-th element of the classifier's output logits $g^q(\boldsymbol{x})$, representing the predicted score for instance $\boldsymbol{x}$ belonging to class $j$. 
Here, $\omega_j$ is the normalized confidence for label $j$, estimated as $\omega_j=\frac{\exp(g^k_j(\boldsymbol{x}))}{\sum_{r \in \mathcal{S}^\prime(j)}\exp(g^k_r(\boldsymbol{x}))}$,
where $\mathcal{S}^\prime(j)=\mathcal{S}$ if $j\in\mathcal{S}$ and $\mathcal{S}^\prime(j)=\bar{\mathcal{S}}$ otherwise.

This construction rewards confident predictions within $\mathcal{S}$ and suppresses overconfident responses in $\bar{\mathcal{S}}$ by normalizing $\omega_j$ within the corresponding set. Concretely, for $j\in\mathcal{S}$ the term reduces to $\psi (g^q_j(\boldsymbol{x}))$ and larger $\omega_j$ places more emphasis on confident candidate labels; for $j\in\bar{\mathcal{S}}$ it becomes $\psi(-g^q_j(\boldsymbol{x}))$ and the penalty scales with $\omega_j$, which more aggressively suppresses semantically confusable non-candidate classes. Because $\omega_j$ is normalized within $\mathcal{S}$ or $\bar{\mathcal{S}}$, the total weighted contribution from each set is controlled (since $\sum_{j\in\mathcal S}\omega_j=1$ and $\sum_{j\in\bar{\mathcal S}}\omega_j=1$), regardless of the size of $\mathcal{S}$. Specifically, as $ \mathcal{S} $ increases, individual weights $\omega_j$ decrease proportionally, ensuring that the aggregate influence from $\mathcal{S}$ and $\bar{\mathcal{S}}$ stays constant. This design not only amplifies discriminative signals for typical ambiguous instances (e.g., disjoint labels with similar features) by widening the confidence gap between candidate and confusable non-candidate classes, but also stabilizes training by preventing gradient magnitudes from fluctuating with varying set sizes. The resulting loss can be cast into the leveraged loss family of \cite{wen2021leveraged}, and thus shares a similar Bayes-consistency intuition (discussed in Appendix~\ref{asec:proof}). In practice, to stabilize training and improve optimization efficiency, we adopt a cross-entropy variant of this loss: $\ell(\boldsymbol{s}_j,\boldsymbol{x})=-\log[(1-\boldsymbol{s}_j)+(2\boldsymbol{s}_j-1)p_j]$, with $p_j=\text{softmax}(g^q(\boldsymbol{x}))_j$ (equivalently, $-\boldsymbol{s}_j\log p_j-(1-\boldsymbol{s}_j)\log(1-p_j)$), which realizes an analogous reward-penalty mechanism in log-probability space.

\subsection{Algorithm Overview}
Given a PLL instance $(\boldsymbol{x}, \mathcal{S})$, the overall loss for training the classification model in CAD is the weighted sum of the confidence adjustment loss $\mathcal{L}_{discls}$ and representation learning loss $\mathcal{L}_{c}$, expressed formally as
\begin{equation}
    \mathcal{L}(\boldsymbol{x}, \mathcal{S}) = \mathcal{L}_{discls}(\boldsymbol{x})+\frac{\beta}{|\mathcal{S}|}\sum_{s\in \mathcal{S}}\mathcal{L}_{c}(\boldsymbol{x}^\prime_s), \label{eq:all}
\end{equation}
where $\beta$ is a weight parameter used to balance the importance of $\mathcal{L}_{discls}$ and $\mathcal{L}_{c}$, $|\mathcal{S}|$ is the size of the label set $\mathcal{S}$, and $\boldsymbol{x}^\prime_s$ represents the label-specific augmentation for $\boldsymbol{x}$. In practice, we adopt vanilla self-distillation as the baseline following DIRK \cite{wu2024distilling}. The overall training process and pseudo-code are outlined in Appendix \ref{asec:pseudocode}.

%% file: sec/4_experiments.tex
\section{Experiments}
\subsection{Experimental Setup}
\noindent\textbf{Datasets.} We evaluated the performance of our proposed method on five widely used datasets: Fashion-MNIST \cite{xiao2017fashion}, CIFAR-10 \cite{krizhevsky2009learning}, CIFAR-100 \cite{krizhevsky2009learning}, Flower \cite{nilsback2008automated}, and Oxford-IIIT Pet \cite{parkhi2012cats}. Following standard settings in ID-PLL \cite{wu2024distilling, xia2022ambiguity}, we constructed partial-label annotations for these datasets. A detailed description of the label generation process and dataset statistics is provided in Appendix~\ref{asec:syndata}. Additionally, we provide the comparison results on the realistic PLCIFAR10 dataset \cite{wang2025realistic} in Appendix~\ref{asec:exp_plcifar10}.

\smallskip\noindent\textbf{Baselines.} We compared CAD with the advanced PLL methods including POP \cite{xu2023progressive}, VALEN \cite{xu2021instance}, IDGP \cite{qiao2022decompositional}, CP-DPLL \cite{wu2022revisiting}, CAVL \cite{zhang2021exploiting}, PICO \cite{wang2022pico}, PRODEN \cite{lv2020progressive}, LWS \cite{wen2021leveraged}, RC \cite{feng2020provably}, CC \cite{feng2020provably}, CEL \cite{yang2025mixed}, ABLE \cite{xia2022ambiguity}, and DIRK \cite{wu2024distilling}. These methods represent a range of strategies, including label refinement, variational inference, and contrastive learning. Appendix~\ref{asec:baselines} provides details of them.

\smallskip\noindent\textbf{Implementations.} We used ResNet18 \cite{he2016identity} as the backbone for Fashion-MNIST \cite{xiao2017fashion} and ResNet34 \cite{he2016identity} for CIFAR-10 \cite{krizhevsky2009learning}, CIFAR-100 \cite{krizhevsky2009learning}, Flower \cite{nilsback2008automated}, and Oxford-IIIT Pet \cite{parkhi2012cats}. The networks were trained using the Stochastic Gradient Descent (SGD) optimizer with an initial learning rate of 0.01, a weight decay of 0.001, and a momentum of 0.9 for 500 epochs, following a cosine learning rate decay strategy. The mini-batch size was set to 64 for Fashion-MNIST, CIFAR-10, and CIFAR-100, and 16 for Flower and Oxford-IIIT Pet. The details of data augmentation and CAD-CAM (introduced in Section \ref{sec:rep}) are provided in Appendix \ref{asec:implementation_detail}.

\subsection{Comparison for Classification}
To evaluate the performance of the proposed method, we report the accuracy of CAD, its variant CAD-CAM (introduced in Section \ref{sec:rep}), and the baselines in Table \ref{tab:accuracy}. The results show that CAD and CAD-CAM achieve the best performance across five benchmarks, including two fine-grained datasets (Flower and Oxford-IIIT Pet). We also observe that CAD-CAM outperforms CAD on fine-grained datasets, as CAD's diffusion-based augmentation relies on class names, which may not accurately capture subtle visual differences—a limitation that suggests future work (discussed in Section \ref{sec:ablation}). Here, all baseline results are cited from DIRK for consistency, while CEL is reproduced under the same settings as it was not reported therein. Despite the strong empirical performance of CAD, its use of edited samples generated by an external diffusion model naturally raises an important question regarding the source of these performance gains:

\smallskip\noindent\textbf{Discussion: Do CAD’s gains rely solely on external generative priors?}
We provide three empirical observations showing that the improvements of CAD do not solely stem from the external diffusion model.
\textbf{First and most importantly}, CAD-CAM—our purely internal instantiation based on CAM-driven feature reweighting—already outperforms all baselines (Table~\ref{tab:accuracy}), demonstrating that the core disentanglement mechanism is effective even without any external generative prior.
\textbf{Second}, directly adding diffusion-edited instances to existing methods (e.g., DIRK, ABLE) and treating the induced labels as supervision degrades their performance (Appendix~\ref{asec:exp_with_aug}), largely due to semantic inconsistency and distribution shift (Appendix~\ref{asec:visual}); this indicates that the gains of CAD do not arise merely from the edited samples, but from how they are structurally integrated into the learning objective.
\textbf{Third}, when these baselines are equipped with our intra-class alignment module—i.e., using the same diffusion-edited instances but within CAD's intra-class regulation—they consistently benefit, confirming that the proposed framework provides a principled way to exploit structured, label-specific augmentations (Appendix~\ref{asec:plug_and_play}).

\subsection{Empirical Analysis of Disentanglement}
To assess CAD's effectiveness in resolving entanglement-induced confusion, we benchmark it against representative methods (RC, ABLE, and DIRK) across three dimensions: (i) accuracy on entangled instances, (ii) inter-class separability via t-SNE and distance measurements, and (iii) class discriminability via confusion matrices.

\smallskip\noindent\textbf{CAD recognizes the entangled instances better.} 
As defined in Section~\ref{sec:problem_setting}, two instances are considered entangled if they: (a) belong to different classes, (b) include each other's ground-truth label in their candidate sets, and (c) have highly similar feature representations. In this experiment, we focus on the most challenging entangled pairs—specifically, the top 0.1\%, 0.01\%, and 0.001\% most similar pairs satisfying conditions (a) and (b). The corresponding accuracy results are shown in Table~\ref{tab:entan_acc}, and full statistics are provided in Appendix~\ref{asec:entan_statistic}.

As shown in Table~\ref{tab:entan_acc}, CAD consistently outperforms all baselines in identifying entangled instances across datasets. Notably, the performance gap is larger on these challenging pairs than in overall accuracy. For instance, on CIFAR-10, CAD surpasses DIRK by 9.28\% on the top 0.001\% entangled pairs, while its overall gain is only 2.70\%. This highlights the critical role of handling entangled instances in improving model performance. Moreover, CAD achieves the largest average Euclidean distance between entangled pairs in 11 out of 15 cases, indicating stronger feature separation. This trend persists even on fine-grained datasets (e.g., Flower, Oxford-IIIT Pet), where CAD maintains robust performance despite increased class similarity. In contrast, contrastive learning-based methods such as ABLE often shrink the distance between entangled pairs (with the smallest distances in 11 of 15 cases), revealing their limitations in distinguishing closely related instances. Notably, as detailed in Appendix \ref{asec:entan_statistic}, some settings involve only one pair of entangled samples. All models misclassify this pair into the same class, leading to identical 50\% accuracy.

\begin{table}[t]
  \centering
  \caption{Accuracy on entangled instances and average Euclidean distance between entangled pairs.}
  \label{tab:entan_acc}
  \resizebox{0.99\linewidth}{!}{
  \begin{tabular}{ccccccccc}
      \toprule
      ~ & ~ & \multicolumn{2}{c}{0.100\%} & \multicolumn{2}{c}{0.010\%} & \multicolumn{2}{c}{0.001\%} \\ 
      \cmidrule(lr){3-4}\cmidrule(lr){5-6}\cmidrule(lr){7-8}
      ~ & ~ & Acc & Distance & Acc & Distance & Acc & Distance \\ 
      \midrule
      \multirow{4}{*}{F-MNIST} & RC & 83.38  & 0.4507  & 74.02  & 0.4036  & 68.75  & 0.3623  \\ 
      ~ & ABLE & 87.75  & 0.0955  & 78.89  & 0.0798  & 72.32  & 0.0865  \\ 
      ~ & DIRK & 89.57  & 0.4890  & 80.91  & 0.4110  & 72.86  & 0.3447  \\ 
      ~ & CAD & \textbf{89.87}  & \textbf{0.5099}  & \textbf{81.39}  & \textbf{0.4372}  & \textbf{74.79}  & \textbf{0.4186}  \\ 
      \midrule
      \multirow{4}{*}{CIFAR-10} & RC & 84.51  & \textbf{0.6161}  & 79.17  & \textbf{0.6088}  & 70.42  & \textbf{0.5588}  \\ 
      ~ & ABLE & 85.65  & 0.1904  & 79.91  & 0.1684  & 70.10  & 0.1415  \\ 
      ~ & DIRK & 91.78  & 0.4247  & 85.88  & 0.4029  & 74.09  & 0.3457  \\ 
      ~ & CAD & \textbf{94.51}  & 0.4714  & \textbf{90.90}  & 0.4315  & \textbf{83.37}  & 0.3880  \\ 
      \midrule
      \multirow{4}{*}{CIFAR-100} & RC & 61.62  & 0.5168  & 55.20  & 0.4726  & 45.19  & 0.4508  \\ 
      ~ & ABLE & 63.45  & 0.1416  & 60.58  & 0.1141  & 50.37  & 0.1253  \\ 
      ~ & DIRK & 70.42  & 0.5461  & 66.61  & 0.4959  & 62.59  & 0.4412  \\ 
      ~ & CAD & \textbf{72.43}  & \textbf{0.5643}  & \textbf{68.80}  & \textbf{0.5233}  & \textbf{67.78}  & \textbf{0.4916}  \\ 
      \midrule
      \multirow{4}{*}{Pet} & RC & 54.20  & 0.4187  & 59.09  & 0.4232  & \textbf{50.00}  & 0.2083  \\ 
      ~ & ABLE & 48.85  & 0.3508  & 63.64  & 0.3218  & \textbf{50.00}  & 0.2185  \\ 
      ~ & DIRK & 63.36  & 0.4064  & 63.64  & 0.4023  & \textbf{50.00}  & \textbf{0.2700}  \\ 
      ~ & CAD & \textbf{72.52}  & \textbf{0.4855}  & \textbf{68.18}  & \textbf{0.4412}  & \textbf{50.00}  & 0.1949  \\ 
      \midrule
      \multirow{4}{*}{Flower} & RC & 62.50  & 0.3563  & \textbf{50.00}  & 0.3736  & \textbf{50.00}  & 0.3736  \\ 
      ~ & ABLE & 62.50  & 0.3688  & \textbf{50.00}  & 0.3542  & \textbf{50.00}  & 0.3542  \\ 
      ~ & DIRK & 50.00  & 0.3229  & \textbf{50.00}  & 0.3229  & \textbf{50.00}  & 0.3229  \\ 
      ~ & CAD & \textbf{75.00}  & \textbf{0.4959}  & \textbf{50.00}  & \textbf{0.5045}  & \textbf{50.00}  & \textbf{0.5045}  \\ 
      \bottomrule
  \end{tabular}
  }
\end{table}

\begin{figure}[t]
  \centering
  \begin{subfigure}[b]{0.24\linewidth}
    \includegraphics[width=\textwidth]{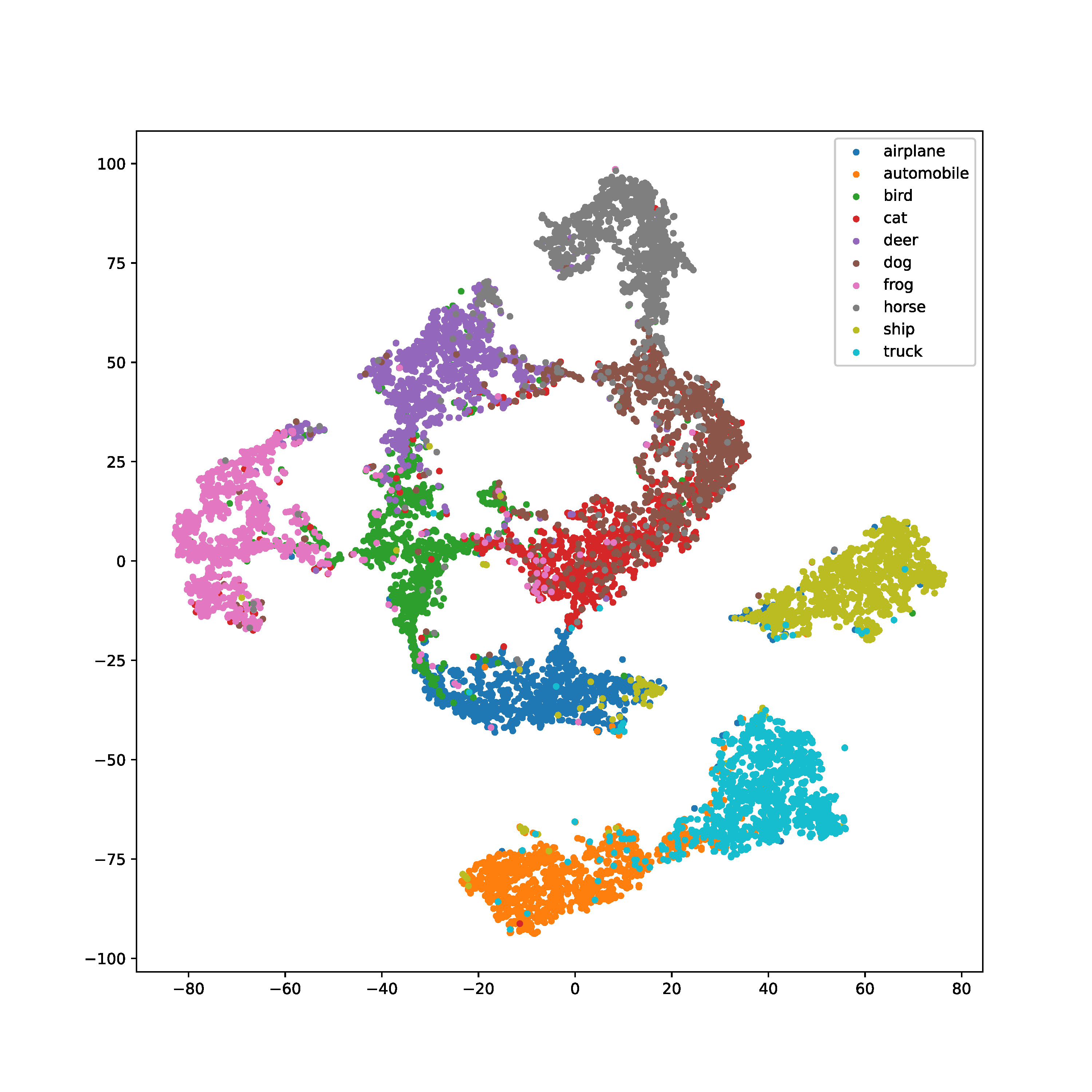}
    \caption{RC}
    \label{fig:TSNE-RC-CIFAR10}
  \end{subfigure}
  \begin{subfigure}[b]{0.24\linewidth}
    \includegraphics[width=\textwidth]{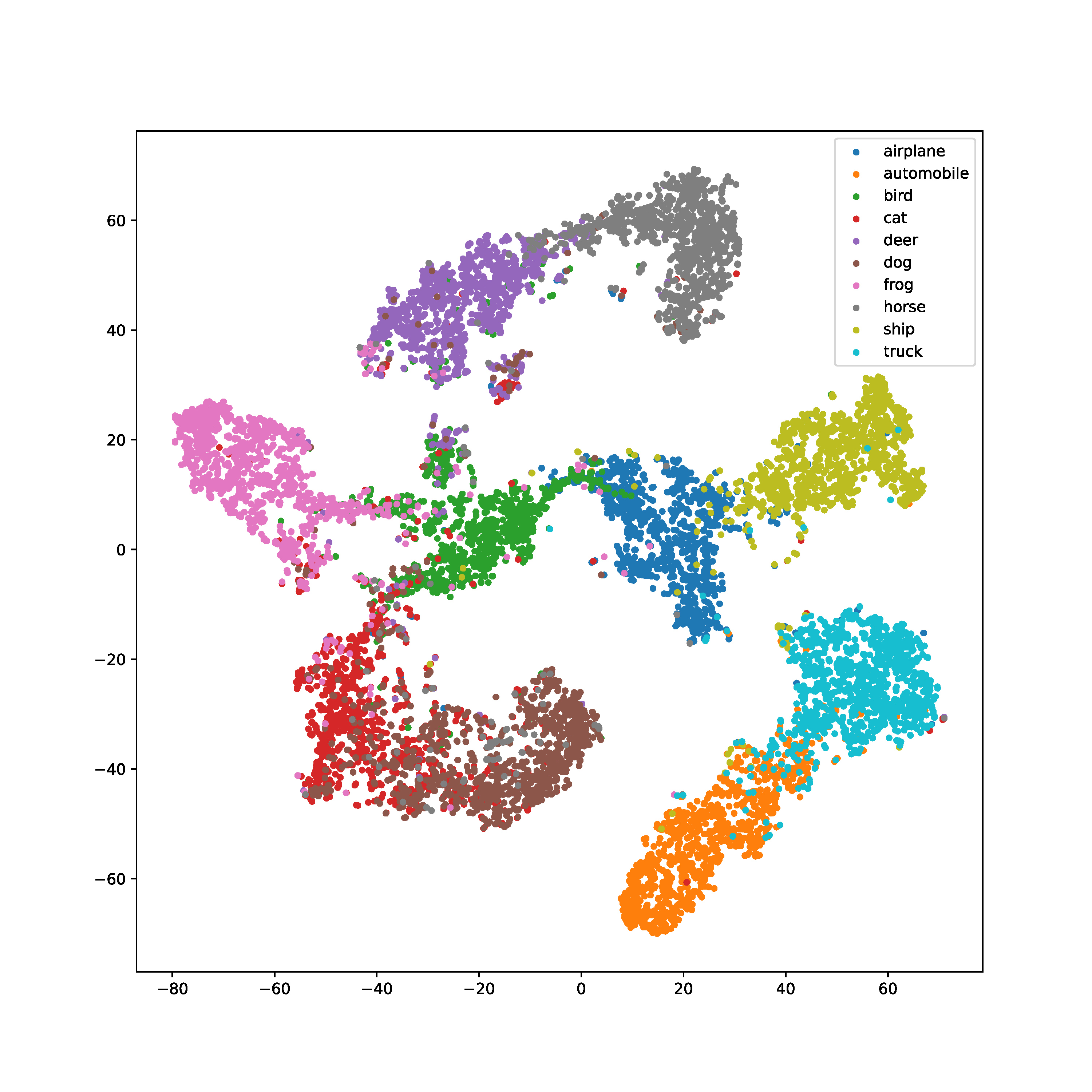}
    \caption{ABLE}
    \label{fig:TSNE-ABLE-CIFAR10}
  \end{subfigure}
  \begin{subfigure}[b]{0.24\linewidth}
    \includegraphics[width=\textwidth]{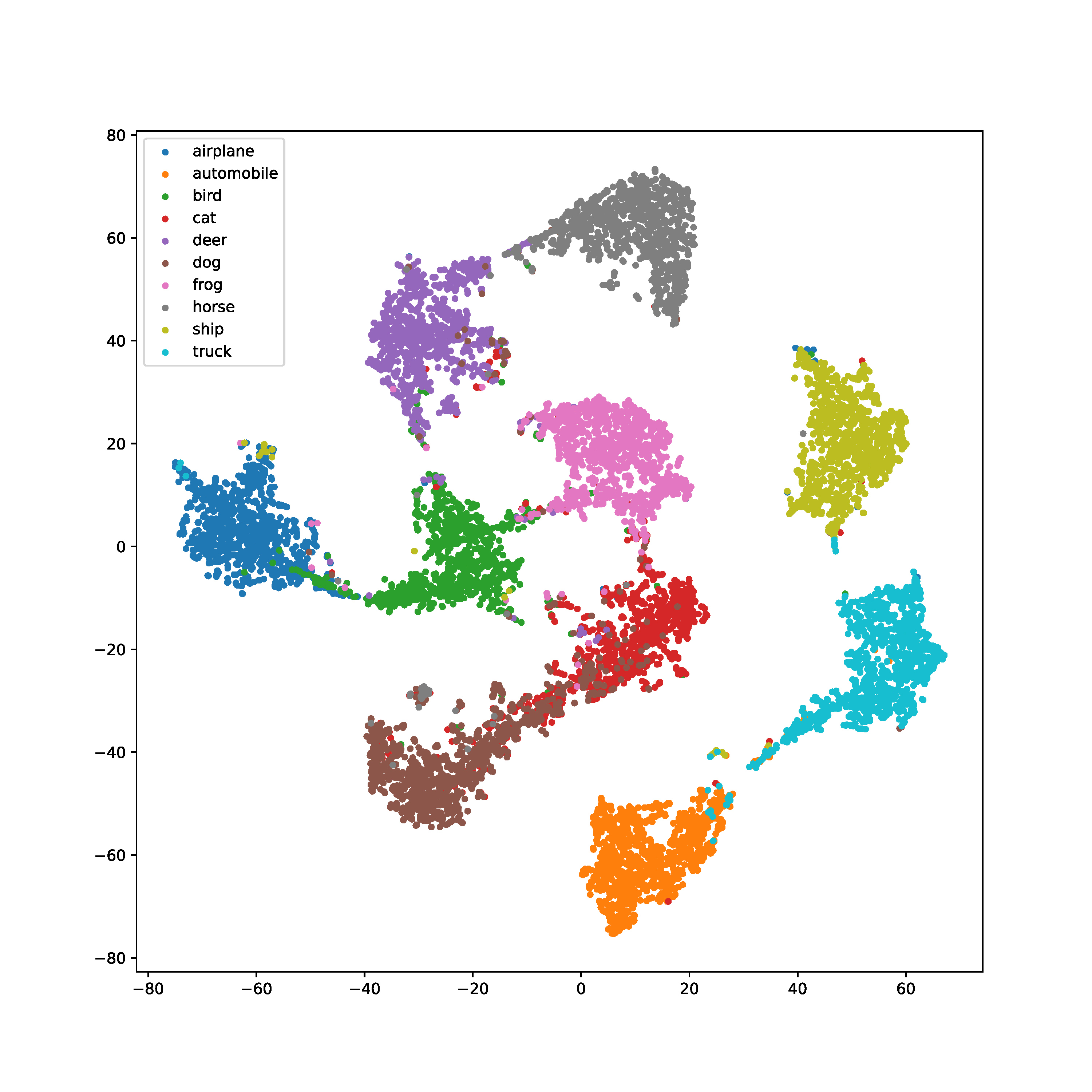}
    \caption{DIRK}
    \label{fig:TSNE-DIRK-CIFAR10}
  \end{subfigure}  
  \begin{subfigure}[b]{0.24\linewidth}
    \includegraphics[width=\textwidth]{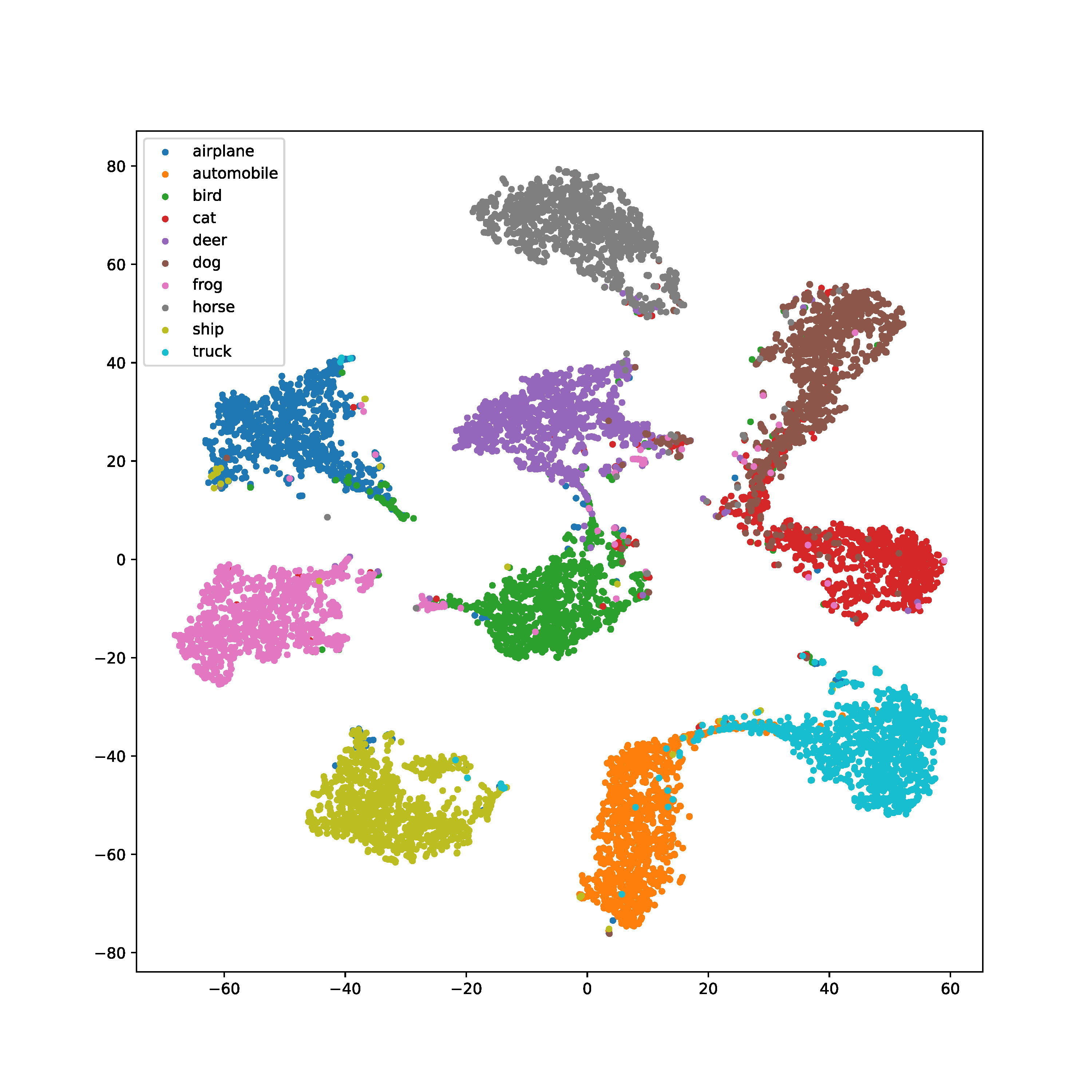}
    \caption{CAD}
    \label{fig:TSNE-DisCon-CIFAR10}
  \end{subfigure}
  \caption{t-SNE visualization on the CIFAR-10 benchmark.}
  \label{fig:t-SNE}
\end{figure}

\begin{figure}[t]
  \centering
  \begin{subfigure}[b]{0.24\linewidth}
    \includegraphics[width=\textwidth]{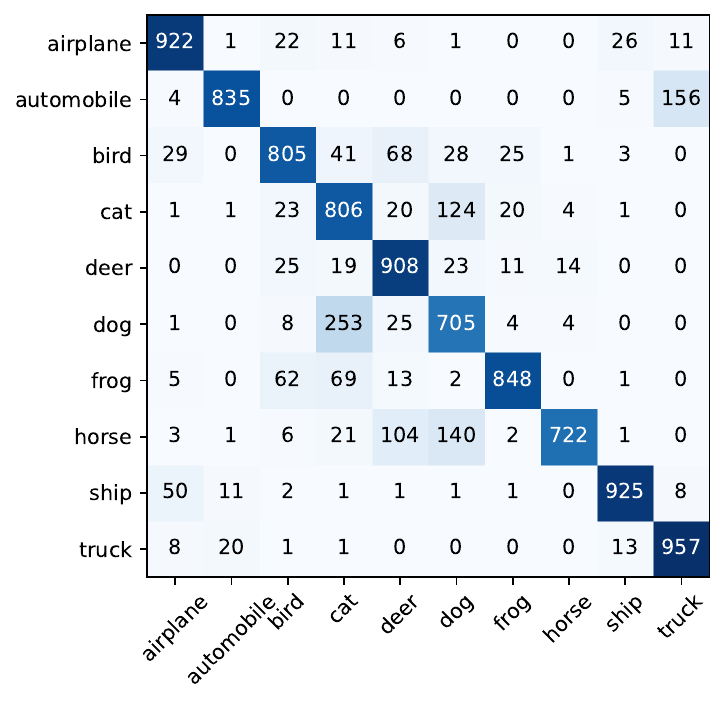}
    \caption{RC}
    \label{fig:CM-RC-CIFAR10}
  \end{subfigure}
  \begin{subfigure}[b]{0.24\linewidth}
    \includegraphics[width=\textwidth]{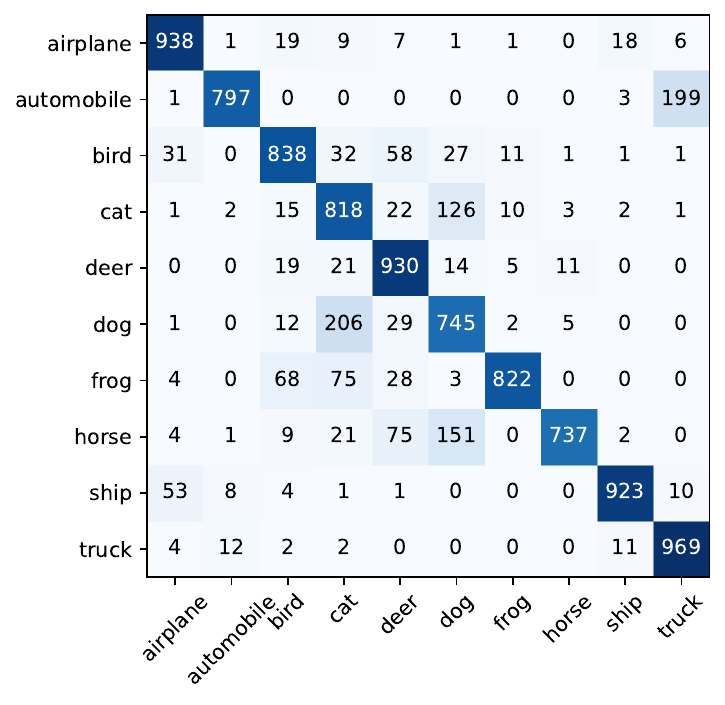}
    \caption{ABLE}
    \label{fig:CM-ABLE-CIFAR10}
  \end{subfigure}
  \begin{subfigure}[b]{0.24\linewidth}
    \includegraphics[width=\textwidth]{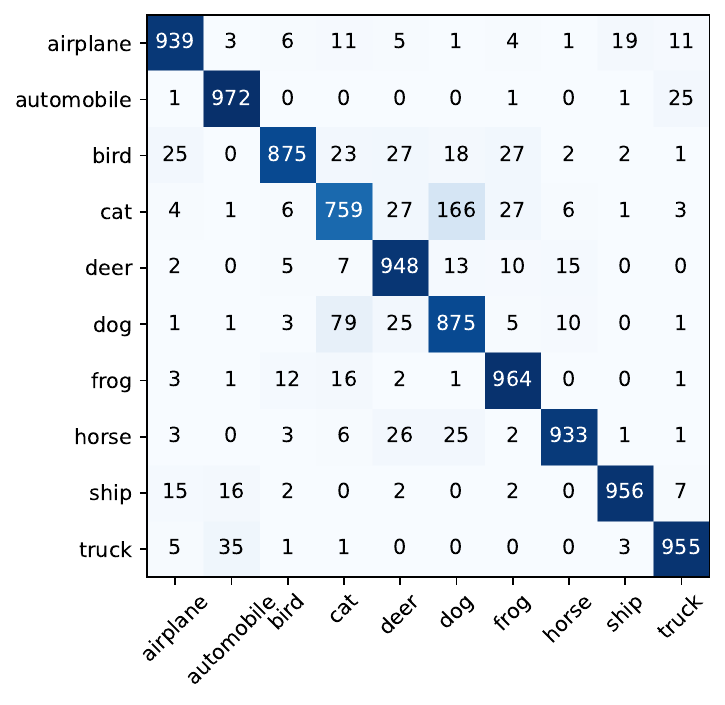}
    \caption{DIRK}
    \label{fig:CM-DIRK-CIFAR10}
  \end{subfigure}  
  \begin{subfigure}[b]{0.24\linewidth}
    \includegraphics[width=\textwidth]{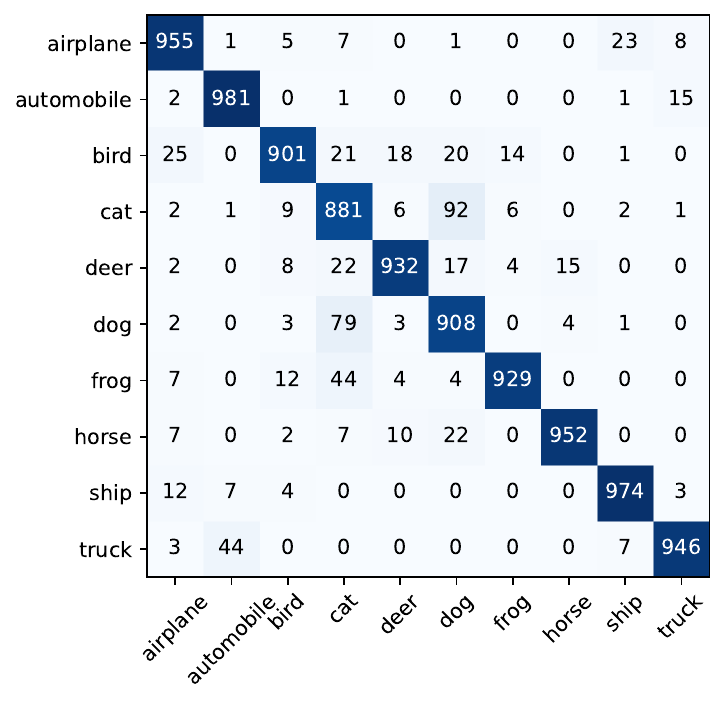}
    \caption{CAD}
    \label{fig:CM-DisCon-CIFAR10}
  \end{subfigure}
  \caption{Confusion matrices on the CIFAR-10 benchmark.}
  \label{fig:CM}
\end{figure}

\smallskip\noindent\textbf{CAD increases the inter-class distances.} 
Figure~\ref{fig:t-SNE} compares the t-SNE \cite{van2008visualizing} visualizations of CAD and representative methods—RC, ABLE, and DIRK—on the CIFAR-10 benchmark. In the figure, RC shows the most severe class overlap, which aligns with its substantially lower accuracy compared to the other methods. Although ABLE and DIRK exhibit reduced mixing compared to RC, the confusion near class boundaries remains apparent. In contrast, CAD produces significantly improved class separability. For example, in challenging categories such as the cat (red) and dog (brown) classes, CAD forms clearer decision boundaries despite occasional misclassifications. Similar results are observed on Fashion-MNIST, as presented in Appendix~\ref{asec:tsne}.

\smallskip\noindent\textbf{CAD reduces confusion between similar classes.} To compare the distinction between confusing classes, we compare confusion matrices and distance metrics for confusing samples in Figure \ref{fig:CM} and Table \ref{tab:distance}, respectively. In Figure \ref{fig:CM}, we compare the confusion matrices on the CIFAR-10 dataset. Based on label overlap (details are reported in Appendix \ref{asec:cm}), the most likely overlapping class pairs are cat-dog and truck-automobile, with 96.62\% of instances from the cat-dog classes holding both cat and dog candidate labels, and 96.25\% for the truck-automobile pair. The results show that CAD reduces both cat-dog and truck-automobile confusion through mitigating entanglement. We additionally report the experiments on the Fashion-MNIST dataset in Appendix \ref{asec:cm}, which show a similar trend.

In Table \ref{tab:distance}, we report the comparison results of similar class distances in the classifier's feature space. We compared ABLE, DIRK, and our CAD on the CIFAR-10 benchmark. Specifically, we evaluated the following: the nearest distance between instances from different classes (Instance), the average nearest-instance distance across pairs of different classes (Avg. Pairwise), and the average distance between class centroids (Centroid). These three distances reflect the proximity of the closest boundary samples for the most similar classes, the closest boundary samples between similar class pairs, and the center distance of the classes. On all three metrics, both CAD and RC show the highest or second-highest values. Compared with RC, CAD leverages intra-class alignment to enhance representation quality, effectively improving classification performance while maintaining inter-class distance.

\begin{table}[t]
  \centering
  \caption{Comparison of distances between similar classes on the CIFAR-10 benchmark (higher is better).}
  \label{tab:distance}
  \begin{tabular}{lcccc}
    \toprule
    Metric & RC & ABLE & DIRK & CAD \\
    \midrule
    Instance         & \underline{0.040} & 0.007 & 0.035 & \textbf{0.057} \\
    Avg. Pairwise    & \textbf{0.414}    & 0.128 & 0.333 & \underline{0.337} \\
    Centroid         & \underline{1.009} & 0.340 & 0.936 & \textbf{1.103} \\
    \bottomrule
  \end{tabular}
\end{table}

\subsection{Ablation Study} \label{sec:ablation}
\smallskip\noindent\textbf{Component ablation.} CAD consists of a representation learning module for intra-class regulation and a confidence adjustment module for inter-class regulation. To evaluate the effectiveness of each component, we perform an ablation study by removing the two modules individually. Table~\ref{tab:abl} reports the results on the CIFAR-10 and Fashion-MNIST benchmarks. Specifically, \textit{w/o RL} denotes the model without the representation learning module, while \textit{w/o CA} refers to the variant where the confidence adjustment module is replaced with a standard weighted loss function. When both modules are removed (\textit{w/o Both}), the loss function of CAD degenerates into a simple weighted loss under a vanilla distillation framework. The results show that both modules contribute positively to performance. In addition, we provide experiments analyzing parameter sensitivity, low-entangled datasets, and non-visual classification datasets in Appendices~\ref{asec:parameter}, \ref{asec:exp_low_entangle} and \ref{asec:exp_text}.

\begin{table}[t]
  \centering
  \caption{Ablation accuracy (\%, mean $\pm$ std).}
  \label{tab:abl}
  \resizebox{0.99\linewidth}{!}{
  \begin{tabular}{lcccc}
    \toprule
    Dataset & CAD & w/o CA & w/o RL & w/o Both \\
    \midrule
    Fashion-MNIST & \textbf{92.14 $\pm$ 0.91} & 91.19 $\pm$ 0.22 & 91.48 $\pm$ 0.61 & 85.30 $\pm$ 0.87 \\
    CIFAR-10      & \textbf{93.57 $\pm$ 0.24} & 93.32 $\pm$ 0.13 & 91.21 $\pm$ 0.52 & 87.81 $\pm$ 1.13 \\
    \bottomrule
  \end{tabular}
  }
\end{table}

\smallskip\noindent\textbf{Fine-grained dataset limitation analysis.} 
As shown in Table~\ref{tab:accuracy}, diffusion-based CAD models underperform relative to CAD-CAM on fine-grained datasets. This is likely because generic image editing tools lack the category-specific semantic priors required for fine-grained recognition. For example, when editing visually similar breeds such as the ``American Pit Bull Terrier'' and the ``American Bulldog'', the model fails to produce sufficiently discriminative features, leading to increased inter-class confusion.

To verify this hypothesis, we manually enriched the editing prompts by providing a detailed visual description for each of the 37 classes in the Oxford-IIIT Pet dataset. For instance, the prompt for American Pit Bull Terrier included attributes such as: \textit{American Pit Bull Terrier: large muscular dog with broad head, strong jaws, short white or white-with-patches coat, loose facial skin, powerful stance.} As shown in Table~\ref{tab:finegrid}, incorporating these cues allows the fine-grained prompting variant of CAD to surpass CAD-CAM. This supports that the performance gap arises from missing fine-grained priors during editing rather than the fundamental limitation of the diffusion model or the CAD framework itself. All prompts are provided in Appendix~\ref{asec:promptcues}.

\begin{table}[t]
  \centering
  \caption{Accuracy (\%) with different prompting.}
  \label{tab:finegrid}
  \begin{tabular}{lc}
  \toprule
  Method & Accuracy (\%) \\
  \midrule
  CAD-CAM & 74.56 ± 3.16 \\
  CAD w/o cues & 69.46 ± 2.35 \\
  CAD with fine-grained prompting & \textbf{76.23 ± 2.17} \\
  \bottomrule
  \end{tabular}
\end{table}

%% file: sec/5_conclusion.tex
\section{Conclusion}

In this paper, we investigate instance-dependent Partial Label Learning (ID-PLL), where class confusion caused by instance entanglement severely degrades classification performance. To address this, we propose the Class-specific Augmentation based Disentanglement (CAD) framework. CAD effectively mitigates entanglement by generating and aligning class-specific augmentations to resolve intra-class misalignment, while simultaneously applying confidence penalties to confusing labels to enhance inter-class separability. Extensive experiments validate the robustness of our approach in mitigating class confusion. Nevertheless, we note a limitation: in specialized fields like medical or industrial imaging, visual semantics are often difficult to verbalize and lie beyond the priors of generic diffusion models, rendering prompt-driven editing infeasible. Although our framework readily accommodates non-diffusion alternatives (e.g., CAD-CAM) to bypass this issue, exploring domain-specific fine-tuning of generative editors under PLL constraints remains a promising direction for future work.

%% file: sec/appendix.tex
\clearpage
\setcounter{page}{1}
\maketitlesupplementary
\setcounter{page}{1}
\appendix
\setcounter{table}{0}
\setcounter{figure}{0}
\setcounter{algorithm}{0}
\setcounter{secnumdepth}{2}
\setcounter{lemma}{0}
\setcounter{theorem}{0}

\section{Methodological and Theoretical Details}
\subsection{Theoretical Analysis of the Classification Loss}
\label{asec:proof}

In Section~3.3, we propose a disambiguation classification loss $\mathcal{L}_{discls}$ to adjust the confidence of candidate and non-candidate labels. This section shows that $\mathcal{L}_{discls}$ can be cast as an instantiation of the Leveraged Weighted (LWS) loss family~\cite{wen2021leveraged}, and thus shares a similar Bayes-consistency intuition under the symmetric-surrogate assumption.

\paragraph{Review of the LWS loss family.}
For a partially labeled instance $(\boldsymbol{x},\mathcal{S})$, a representative LWS-form loss takes the leveraged weighted form:
\begin{equation}
\begin{aligned}
\tilde{\mathcal{L}}_{\mathrm{LWS}}(\mathcal{S}, g(\boldsymbol{x}))
&= \sum_{j\in\mathcal{S}} w_j\,\psi(g_j(\boldsymbol{x}))\\
&+ \beta \sum_{j\notin\mathcal{S}} w_j\,\psi(-g_j(\boldsymbol{x})),
\end{aligned}
\label{eq:lws_general}
\end{equation}
where $w_j\ge 0$ are label-wise weights, $\beta>0$ balances the penalty on non-candidate labels, and $\psi(\cdot)$ is a non-increasing binary surrogate satisfying the symmetric condition $\psi(t)+\psi(-t)=1$. Prior work provides a Bayes-consistency analysis for such leveraged weighted losses under symmetric surrogates and specific candidate-set generation assumptions~\cite{wen2021leveraged}.

\paragraph{Formulation mapping.}
Recall our proposed loss:
\begin{equation}
\mathcal{L}_{discls}(\boldsymbol{x})=\sum_{j\in\mathcal{Y}}\omega_j\,\ell(s_j,\boldsymbol{x}), 
\end{equation}
where $s_j=\mathbb{I}[j\in\mathcal{S}]$ is the indicator function denoting whether label $j$ is in the candidate set $\mathcal{S}$. Under a symmetric surrogate $\psi(\cdot)$ satisfying $\psi(t)+\psi(-t)=1$, our per-label term can be written as:
\begin{equation}
\ell(s_j,\boldsymbol{x})= s_j\,\psi(g^q_j(\boldsymbol{x})) + (1-s_j)\,\psi(-g^q_j(\boldsymbol{x})),
\end{equation}
which yields:
\begin{equation}
\begin{aligned}
\mathcal{L}_{discls}(\boldsymbol{x})
&= \sum_{j\in\mathcal{Y}} \omega_j \Big[ s_j\,\psi(g^q_j(\boldsymbol{x})) + (1-s_j)\,\psi(-g^q_j(\boldsymbol{x})) \Big] \\
&= \sum_{j\in\mathcal{S}} \omega_j \Big[ 1 \cdot \psi(g^q_j(\boldsymbol{x})) + 0 \cdot \psi(-g^q_j(\boldsymbol{x})) \Big] \\
&\quad + \sum_{j\notin\mathcal{S}} \omega_j \Big[ 0 \cdot \psi(g^q_j(\boldsymbol{x})) + 1 \cdot \psi(-g^q_j(\boldsymbol{x})) \Big] \\
&= \sum_{j\in\mathcal{S}} \omega_j \psi(g^q_j(\boldsymbol{x})) + \sum_{j\notin\mathcal{S}} \omega_j \psi(-g^q_j(\boldsymbol{x})).
\end{aligned}
\end{equation}
Therefore, Eq.~\eqref{eq:lws_general} reduces to our formulation by setting the leverage parameter $\beta=1$ and matching the weights $w_j=\omega_j$. Since our $\mathcal{L}_{discls}$ admits the leveraged weighted form in Eq.~\eqref{eq:lws_general} under a symmetric surrogate, it naturally aligns with the Bayes-consistency intuition developed for LWS losses in~\cite{wen2021leveraged}.

In practice, we optimize a cross-entropy variant for numerical stability, which preserves the same reward--penalty mechanism in the log-probability space. Notably, this is consistent with the empirical observations in~\cite{wen2021leveraged}, which demonstrate that the cross-entropy loss provides greater stability during the optimization process.

\subsection{Pseudo Code of Training}
\label{asec:pseudocode}
The complete training process is outlined in Algorithm \ref{alg:algorithm}. First, CAD generates class-specific augmentations for each instance based on the candidate labels and computes the contrastive loss from these augmentations (steps 1-2). Next, it calculates the disambiguation classification loss for the instances to obtain the overall loss and updates the model (steps 3-5). Meanwhile, two queues are maintained to store keys and their corresponding confidences, which are updated at each batch (step 6).
\begin{algorithm}[t]
    \caption{Training framework with mini-batch}
    \label{alg:algorithm}
    \begin{algorithmic}[1]
        \REQUIRE Partial label training batch $\mathcal{B}$, augmentations generation model $E$, query model $e_q$ and classifier model $g_q$ with a shared backbone, key model $e_k$ and confidence estimator model $g_k$ with a shared backbone.\\
        \ENSURE Optimized classifier $f(\boldsymbol{x})=\arg \max_j g^q_j(\boldsymbol{x})$.
        \STATE Generate class-specific augmentations $\boldsymbol{x}^\prime_s = E(\boldsymbol{x},I(s))$ for every $(\boldsymbol{x}, \mathcal{S}) \in \mathcal{B}$ and for each $s \in \mathcal{S}$.
        \STATE Calculate the contrastive loss $\mathcal{L}_{c}(\boldsymbol{x}^\prime_s)$ for augmentations by Eq. (\ref{eq:lcon}) of the main paper.
        \STATE Calculate the disambiguation classification loss $\mathcal{L}_{discls}(\boldsymbol{x})$ for instance by Eq. (\ref{eq:dc}) of the main paper.
        \STATE Calculate the final loss $\mathcal{L}(\boldsymbol{x}, \mathcal{S})$ according to Eq. (\ref{eq:all}) and update the network $e_q$ and $g_q$.
        \STATE momentum update $e_k$ and $g_k$ using $e_q$ and $g_q$.
        \STATE Update the key queue and the confidence queue.
    \end{algorithmic}
\end{algorithm}

\section{Dataset Details}
\label{asec:syndata}
\subsection{Dataset Synthesis}
In the main paper, following common practice \cite{xu2021instance,xia2022ambiguity,wu2024distilling}, we synthesized some partial label datasets. Specifically, a model trained on the clean dataset is used as the annotator to predict the class posterior probabilities of the instances. Then, the top few labels with the highest posterior probabilities are selected as candidate labels, with the number of candidate labels controlled by a hyperparameter $\tau$. The overall algorithmic is illustrated in Algorithm \ref{alg:data}.

\begin{algorithm}[ht]
    \caption{Pseudocode for data synthesis}
    \label{alg:data}
    \begin{algorithmic}[1]
        \REQUIRE Dataset $\{(\boldsymbol{x}_i,y_i)\}_{i=1}^n$, annotator model $f: \mathcal{X}\rightarrow \mathbb{R}^C$ that maps instances sampled from feature space $\mathcal{X}$ to a $C$-dimensional class posterior space, hyperparameter $\tau$.\\
        \ENSURE The set of candidate labels $\mathcal{S}_i$ for each instance $x_i$ where $i \in \{1, \dots, n\}$.
        \STATE For all $\boldsymbol{x}_i$, estimate its class posterior probability $\boldsymbol{p}_i = f(\boldsymbol{x}_i)$, where $\boldsymbol{p}_{ij}$ represents the posterior estimate $\hat{P}(j|\boldsymbol{x
        }_i)$ for class $j$.
        \STATE Compute the normalized class posterior probability as $\boldsymbol{p}^\prime_{ij} = \frac{\boldsymbol{p}_{ij}}{\max_{j \neq y_i} \boldsymbol{p}_{ij}}$.
        \STATE Calculate the label flip probability as $\boldsymbol{p}^{flip}_{ij} = \frac{\boldsymbol{p}^\prime_{ij}(C-1)}{\sum_{j \neq y_i} \boldsymbol{p}^\prime_{ij}}\times \tau$. Subsequently, if $\boldsymbol{p}^{flip}_{ij} > 1$, set $\boldsymbol{p}^{flip}_{ij} = 1$.
        \STATE Sample from a binomial distribution with probability $\boldsymbol{p}^{flip}_{ij}$. If the result is 1, add the $j$-th label to the candidate label set $\mathcal{S}_i$ for instance $\boldsymbol{x}_i$.
    \end{algorithmic}
\end{algorithm}

\subsection{Dataset Statistics}
The statistics of the datasets used in the experiments are summarized in Table \ref{tab:statistics}.
\begin{table}[ht]
    \centering
    \caption{The statistics of datasets. Here, labels refers to the average number of candidate labels assigned to each instance.}
    \label{tab:statistics}
    \resizebox{0.99\linewidth}{!}{
    \begin{tabular}{cccccc}
        \toprule
        Dataset & Train & Test & Dims & Classes & Labels\\ 
        \midrule
        Fashion-MNIST & 60,000 & 10,000 & $28 \times 28$ & 10 & 7.40\\
        CIFAR-10 & 50,000 & 10,000 & $32 \times 32$ & 10 & 5.89 \\ 
        CIFAR-100 & 50,000 & 10,000 & $32 \times 32$ & 100 & 9.40 \\ 
        Flower & 1,020 & 6,149 & $224 \times 224$ & 102 & 5.49 \\ 
        Oxford-IIIT Pet & 3,680 & 3,669 & $224 \times 224$ & 37 & 3.73 \\ 
        \bottomrule
    \end{tabular}
  }
\end{table}

\subsection{Statistic of Entangled Instances} \label{asec:entan_statistic}
In this section, we provide the statistic and thresholds of entangled instance pairs with the highest similarity (top 0.1\%, 0.01\%, and 0.001\%) in Table \ref{tab:entan_statistic}.

\begin{table}[t]
  \centering
  \caption{Statistic of entangled instances at different thresholds. \#Pairs and \#Instances represent the number of entangled pairs and instances, respectively. Since each instance can be entangled with multiple other instances, the number of pairs and instances may not follow a strict ratio.}
  \label{tab:entan_statistic}
  \resizebox{0.99\linewidth}{!}{
  \begin{tabular}{llllllllll}
      \toprule
      Ratio & \multicolumn{3}{c}{0.100\%} & \multicolumn{3}{c}{0.010\%} & \multicolumn{3}{c}{0.001\%} \\ 
      \cmidrule(lr){2-4}\cmidrule(lr){5-7}\cmidrule(lr){8-10}
      ~ & \#Pairs & \#instances & $\xi$ & \#Pairs & \#instances & $\xi$ & \#Pairs & \#instances & $\xi$  \\ 
      \midrule
      Fashion-MNIST & 822,911 & 42,851 & 0.8774  & 82,291 & 13,345 & 0.9594  & 8,229 & 3,114 & 0.9859   \\ 
      CIFAR-10 & 374,916 & 47,655 & 0.8352  & 37,491 & 17,513 & 0.9088  & 3,749 & 2,833 & 0.9478   \\ 
      CIFAR-100 & 23,041 & 10,007 & 0.8303  & 2,304 & 1,644 & 0.8,971  & 230 & 270 & 0.9364   \\ 
      Oxford-IIIT & 133 & 131 & 0.8972  & 13 & 22 & 0.9211  & 1 & 2 & 0.9407   \\ 
      Flower & 5 & 8 & 0.7740  & 1 & 2 & 0.8137  & 1 & 2 & 0.8137  \\
      \bottomrule
  \end{tabular}
  }
\end{table}

\section{Baselines and Implementation}
\subsection{Details of Baselines}
\label{asec:baselines}
We compared our method with the advanced PLL methods including POP \cite{xu2023progressive}, VALEN \cite{xu2021instance}, IDGP \cite{qiao2022decompositional}, CP-DPLL \cite{wu2022revisiting}, CAVL \cite{zhang2021exploiting}, PICO \cite{wang2022pico}, PRODEN \cite{lv2020progressive}, LWS \cite{wen2021leveraged}, RC \cite{feng2020provably}, CC \cite{feng2020provably}, ABLE \cite{xia2022ambiguity}, and DIRK \cite{wu2024distilling}. These methods represent a range of strategies, including label refinement, variational inference, knowledge distillation, and contrastive learning. We compared our method with the following advanced PLL methods: POP \cite{xu2023progressive}: a label refinement algorithm that iteratively updates both the classifier and the labels; VALEN \cite{xu2021instance}: an algorithm that uses variational inference to estimate the latent label distribution; IDGP \cite{qiao2022decompositional}: a framework utilizes separate categorical and Bernoulli models to capture instance-specific label dependencies and maximizes posterior probabilities; CP-DPLL \cite{wu2022revisiting}: a framework that integrates supervised learning on non-candidate labels with consistency regularization on candidates; CAVL \cite{zhang2021exploiting}: a label disambiguation algorithm that adopts the class activation map mechanism for label selection to obtain potentially reliable labels; PICO \cite{wang2022pico}: a framework that uses contrastive learning to help label disambiguation; PRODEN \cite{lv2020progressive}: a self-training framework that incrementally refines label assignments; LWS \cite{wen2021leveraged}: a loss weighting strategy that balances the losses of candidate labels and non-candidate labels; RC \cite{feng2020provably}: a label weighting algorithm designed to construct risk-consistent loss; CC \cite{feng2020provably}: a label weighting algorithm designed to construct classifier-consistent loss; CEL \cite{yang2025mixed}: a class-wise embedding framework that uses associative and prototype losses to explore intra-sample relationships; ABLE \cite{xia2022ambiguity}: a framework that leverages label ambiguity to construct a contrastive learning setup, aiding in representation learning for PLL; DIRK \cite{wu2024distilling}: an algorithm that learns reliable label distributions by rectifying the relative confidence of candidate and non-candidate labels.

\subsection{Additional Implementation Details}\label{asec:implementation_detail}
This section provides further implementation details that complement the descriptions in the main paper. All experiments in this paper are implemented using the PyTorch framework. For datasets including Fashion-MNIST, CIFAR-10, CIFAR-100, and PLCIFAR10, training was conducted on a single NVIDIA RTX 4090 GPU. For Oxford Flowers and Oxford-IIIT Pet, we used a single NVIDIA A100 GPU due to their higher resolution and increased memory requirements. To ensure fair comparison, all baseline methods are implemented using a unified backbone and the same data augmentation pipelines \cite{wu2024distilling}. For hyperparameters of CAD, we empirically set default $\beta = 1$, $\tau = 0.12$, and $\tau_2 = 0.4$ following \cite{wu2024distilling}. Below, we provide additional details on the implementation of the CAD-CAM method and the data augmentation strategies used in our framework. 

\subsection{Implementation Details of CAD-CAM}\label{asec:cadcam}
To implement the CAM-based augmentation required by CAD-CAM, we first pretrain the model for 50 epochs using a weighted PLL cross-entropy loss. After this warm-up phase, we apply the proposed augmentation strategy: for each candidate label in a sample, we compute the corresponding class activation map (CAM) and generate class-specific augmented samples accordingly. Specifically, we compute standard CAMs as a channel-weighted sum of feature maps followed by ReLU, then normalize and resize them to the input resolution; we retain the top 30\% activations to form class-specific masks. Once the augmented samples are obtained, the contrastive loss is introduced into the training objective. These class-specific augmentations are refreshed every 50 epochs using the updated model predictions. If a CAM is nearly uniform—i.e., no regions are strongly activated—the corresponding class-specific augmentation is discarded for that class. To mitigate potential artifacts or sharp transitions introduced during this augmentation process, we further apply a Gaussian blur with a kernel size of 5 and a standard deviation of 1.0 to each augmented image.

In our implementation, we empirically set the contrastive loss weight to $\beta = 2.0$ and the temperature parameter to $\tau_2 = 0.1$. And to mitigate potential artifacts or sharp transitions introduced during the augmentation process, we apply a Gaussian blur transformation with a kernel size of 5 and a standard deviation of 1.0 to each augmented image.  All other hyperparameters are kept consistent with the original CAD setting.

\subsection{Data Augmentation Details}\label{asec:aug}
We adopt three data augmentation strategies in our framework: a key view and a query view for contrastive learning, and a disambiguation view for label disambiguation. For class-specific augmentation, we employ both CAM-based and class-specific diffusion–based methods. Following DIRK~\cite{wu2024distilling}, class-specific augmented samples are further transformed to generate the key and query views, which are used to compute the contrastive loss. The original samples are augmented into disambiguation views for computing the PLL classification loss with confidence adjustment. For CIFAR-10, PLCIFAR10, and CIFAR-100, the key view is generated via random flipping, resizing, cropping, color jittering, and random grayscaling; the query view uses flipping, resizing, cropping, and RandAugment~\cite{cubuk2020randaugment}; and the disambiguation view is created using flipping, cropping, padding, Cutout~\cite{devries2017improved}, and AutoAugment~\cite{cubuk2018autoaugment}. For Fashion-MNIST, the disambiguation view includes horizontal flipping, reflective cropping, Cutout, and grayscale conversion; the key view uses resized cropping and flipping; and the query view further includes ColorJitter and random grayscaling. For Oxford Flowers and Pets, we apply RandomResizedCrop, flipping, and ImageNetPolicy for the disambiguation view; the key view incorporates ColorJitter and grayscale; and the query view adopts RandomAugment. To ensure fair comparison, we apply the same augmentation pipeline across all methods. Moreover, all baselines are implemented within a unified knowledge distillation framework following DIRK.

\section{Additional Experiments}

\subsection{Comparison of Training Cost} \label{asec:training_cost}
While our main instantiation of CAD leverages a pre-trained diffusion model for class-specific augmentation, this component belongs to the framework rather than the core objective and can be replaced by lighter alternatives. On CIFAR-10, diffusion-based editing takes 5.91 GPU-hours on a single RTX 4090, accounting for about 24\% of CAD’s total cost of 24.66 GPU-hours (Table~\ref{tab:gpu_hours}). Since the diffusion stage is performed offline and is highly parallelizable, its wall-clock time can be further reduced by distributing it across multiple GPUs. Moreover, when efficiency is a primary concern, CAD can be instantiated with the CAM-based variant CAD-CAM, which avoids diffusion entirely and reduces the total cost to 18.50 GPU-hours on CIFAR-10 and 25.23 GPU-hours on CIFAR-100.

Compared with existing ID-PLL methods, the additional GPU-hours of CAD and CAD-CAM mainly arise from the increased number of class-specific augmentations participating in training. The training costs of several representative methods are summarized in Table~\ref{tab:gpu_hours}. Despite this extra GPU-time overhead, CAD and its lightweight variant CAD-CAM achieve substantial performance improvements over these baselines.

\begin{table}[ht]
\centering
\caption{GPU-hours on CIFAR-10 and CIFAR-100 benchmarks.}
\label{tab:gpu_hours}
\begin{tabular}{lcc}
  \toprule
  Method & CIFAR-10 & CIFAR-100 \\
  \midrule
  DIRK & 10.19 & 14.37 \\
  ABLE & 13.89 & 19.15 \\
  CAD-CAM & 18.50 & 25.23 \\
  CAD & 24.66 & 35.54 \\
  \bottomrule
\end{tabular}
\end{table}

\subsection{Experiments with Diffusion-based Augmentation}
\label{asec:exp_with_aug}

This section compares the performance of DIRK, ABLE, and RC with their augmentations using the same diffusion-augmented data as CAD (denoted as DIRK-C, ABLE-C, and RC-C) on the Fashion-MNIST, CIFAR-10, and CIFAR-100 benchmarks. All methods are evaluated under identical experimental conditions: they use the same original training data combined with the same class-specific diffusion-augmented data, employ identical input transformations, and are implemented within the same backbone.

As shown in Table \ref{tab:diff_acc}, the accuracy of other baseline methods decreases after incorporating the augmentation data. This supports the analysis in Section \ref{asec:visual}, where directly using such data with the label induced them introduces more noise, ultimately affecting classification performance. \textbf{It is worth noting that this is why these data were not used for other baseline comparisons in Table \ref{tab:accuracy}}, despite it seemingly being a fairer approach. In contrast, CAD effectively extracts useful signals by aligning same-class enhanced features, which is plug-and-play and consistently boosts baselines (Section \ref{asec:plug_and_play}).

\begin{table*}[ht]
  \centering
  \caption{Comparison of accuracy (expressed in percentage) with Diffusion-based Augmentation. The best results for each benchmark are highlighted in \textbf{bold}.}
  \label{tab:diff_acc}
  \setlength{\tabcolsep}{14pt}
  \begin{tabular}{cccc}
      \toprule
        & Fashion-MNIST & CIFAR-10 & CIFAR-100 \\ 
      \midrule
       RC-C & 66.66 ± 0.99 & 76.80 ± 0.27 & 59.63 ± 0.20 \\ 
       RC & 84.87 ± 1.48 & 87.53 ± 0.94 & 65.26 ± 0.40 \\ 
      \midrule
       ABLE-C & 78.41 ± 0.35 & 59.66 ± 1.04 & 58.03 ± 0.28 \\ 
       ABLE & 89.81 ± 0.08 & 83.92 ± 0.67 & 63.92 ± 0.39 \\ 
      \midrule
       DIRK-C & 79.05 ± 0.49 & 85.06 ± 0.15 & 62.66 ± 0.18 \\ 
       DIRK & 91.48 ± 0.21 & 91.48 ± 0.21 & 68.77 ± 0.49 \\ 
      \midrule
       CAD & \textbf{92.14 ± 0.91} & \textbf{93.57 ± 0.24} & \textbf{72.03 ± 1.88} \\ 
      \bottomrule
  \end{tabular}
\end{table*}

\subsection{Plug-and-Play Integration with Existing Methods} \label{asec:plug_and_play}
To evaluate the generality of our approach, we plug CAD into several representative ID-PLL methods, including POP, ABLE, and DIRK. As shown in Table \ref{tab:plug_and_play}, CAD consistently improves all baselines across both CIFAR-10 and CIFAR-100. Notably, even strong methods such as DIRK benefit from CAD, achieving gains of 1.99\% and 3.58\% on CIFAR-10 and CIFAR-100, respectively. These results demonstrate that CAD serves as a flexible and effective plug-and-play module that can be seamlessly integrated into existing frameworks to enhance instance disentanglement and boost classification performance.

\begin{table}[ht]
\centering
\caption{Accuracy (\%) comparison on CIFAR-10 and CIFAR-100. ``+CAD'' denotes equipping the baseline with our proposed CAD contrastive learning module.}
\label{tab:plug_and_play}
\begin{tabular}{lcc}
\toprule
Method & CIFAR-10 & CIFAR-100 \\
\midrule
POP & 89.55 ± 0.36 & 64.57 ± 0.37 \\
POP+CAD & \textbf{91.56 ± 0.80} & \textbf{66.56 ± 2.86} \\
\midrule
ABLE & 83.92 ± 0.67 & 63.92 ± 0.39 \\
ABLE+CAD & \textbf{85.21 ± 4.08} & \textbf{68.63 ± 1.13} \\
\midrule
DIRK & 90.87 ± 0.25 & 68.77 ± 0.49 \\
DIRK+CAD & \textbf{92.86 ± 0.17} & \textbf{72.35 ± 0.53} \\
\bottomrule
\end{tabular}
\end{table}

\subsection{Experiments on PLCIFAR10} \label{asec:exp_plcifar10}
To evaluate the real-world applicability of CAD, we additionally tested it on the PLCIFAR10 dataset, which contains partial labels derived from human annotations. As shown in Table {\ref{tab:plcifar10_results}}, CAD outperforms the competitive baselines DIRK and ABLE. These results demonstrate CAD's ability to generalize to real-world scenarios.

\begin{table}[ht]
  \centering
  \caption{Accuracy (\%) on the PLCIFAR10 dataset with partial labels from human annotations.}
  \label{tab:plcifar10_results}
  \begin{tabular}{lc}
  \toprule
  Method & Accuracy (\%) \\
  \midrule
  ABLE   & 91.82 \\
  DIRK   & 92.91 \\
  CAD (Ours)   & \textbf{93.24} \\
  \bottomrule
\end{tabular}
\end{table}

\subsection{Experiments for Confusion Matrix} 
\label{asec:cm}
\begin{figure}[ht]
  \centering
  \begin{subfigure}[b]{0.49\linewidth}
    \includegraphics[width=\textwidth]{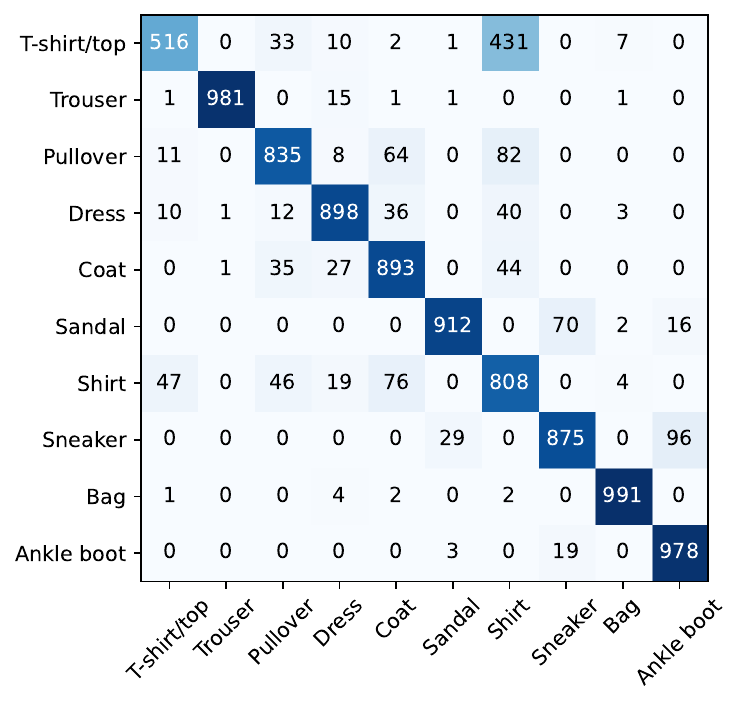}
    \caption{Confusion Matrix for RC}
    \label{fig:CM-RC-fmnist}
  \end{subfigure}
  \begin{subfigure}[b]{0.49\linewidth}
    \includegraphics[width=\textwidth]{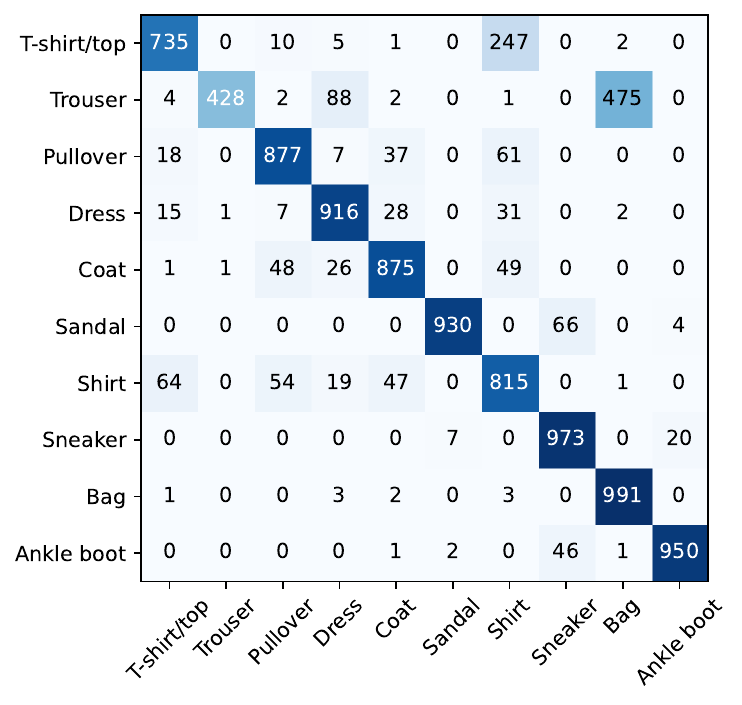}
    \caption{Confusion Matrix for ABLE}
    \label{fig:CM-ABLE-fmnist}
  \end{subfigure}
  
  \begin{subfigure}[b]{0.49\linewidth}
    \includegraphics[width=\textwidth]{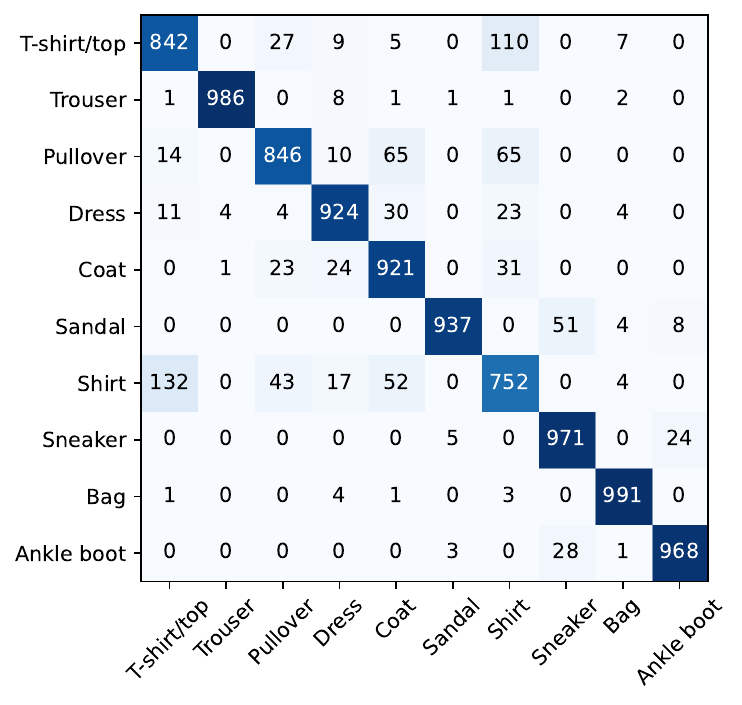}
    \caption{Confusion Matrix for DIRK}
    \label{fig:CM-DIRK-fmnist}
  \end{subfigure}  
  \begin{subfigure}[b]{0.49\linewidth}
    \includegraphics[width=\textwidth]{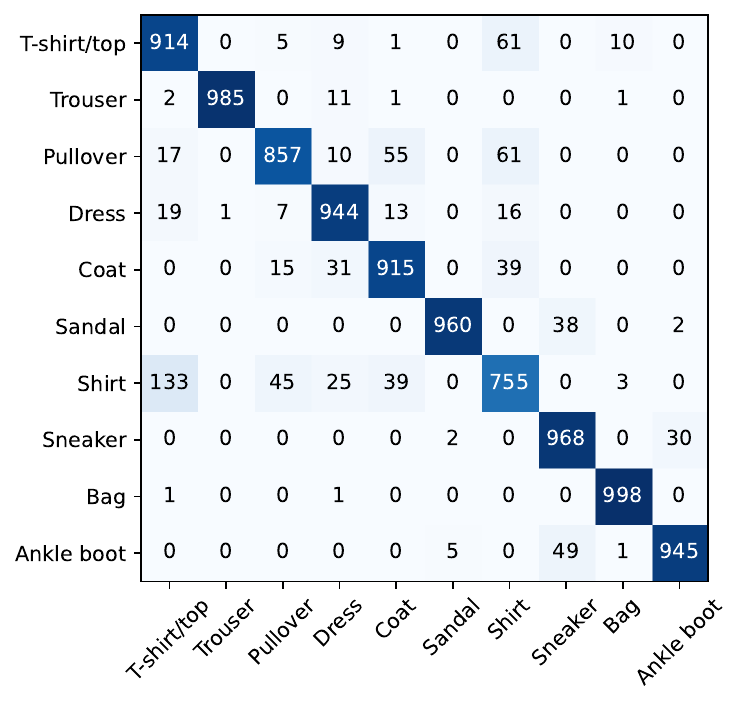}
    \caption{Confusion Matrix for CAD}
    \label{fig:CM-DisCon-fmnist}
  \end{subfigure}
  \caption{Confusion Matrix on the Fashion-MNIST benchmark.}
  \label{fig:CM-fmnist}
\end{figure}

\begin{figure}[ht]
  \centering
  \begin{subfigure}[b]{0.49\linewidth}
    \includegraphics[width=\textwidth]{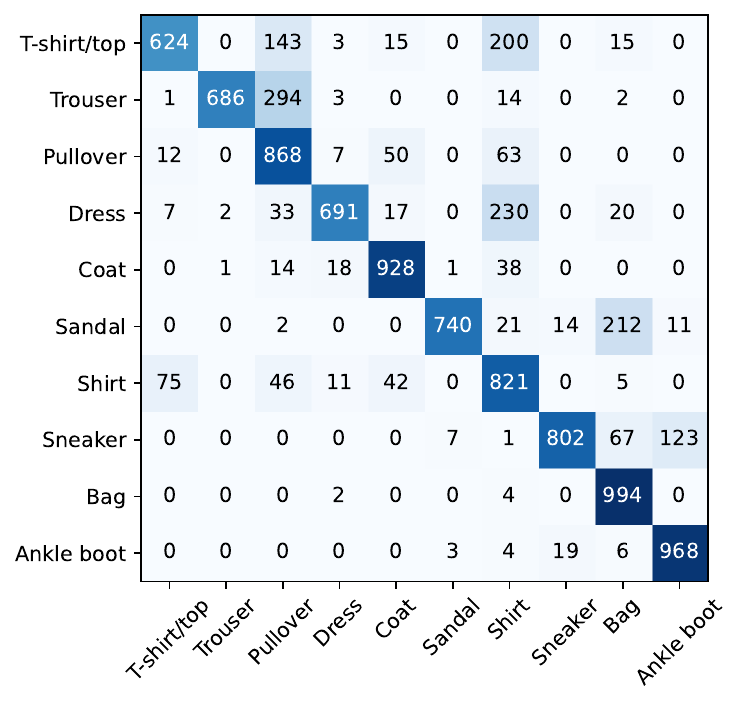}
    \caption{Without Both}
    \label{fig:CM-fmnist-woBoth}
  \end{subfigure}
  \begin{subfigure}[b]{0.49\linewidth}
    \includegraphics[width=\textwidth]{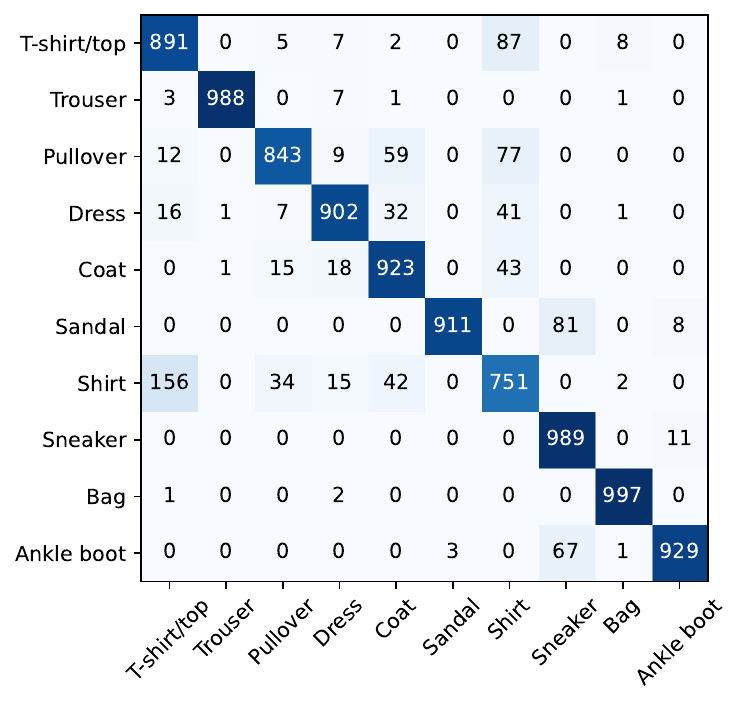}
    \caption{Without LD}
    \label{fig:CM-fmnist-woLD}
  \end{subfigure}  
  \begin{subfigure}[b]{0.49\linewidth}
    \includegraphics[width=\textwidth]{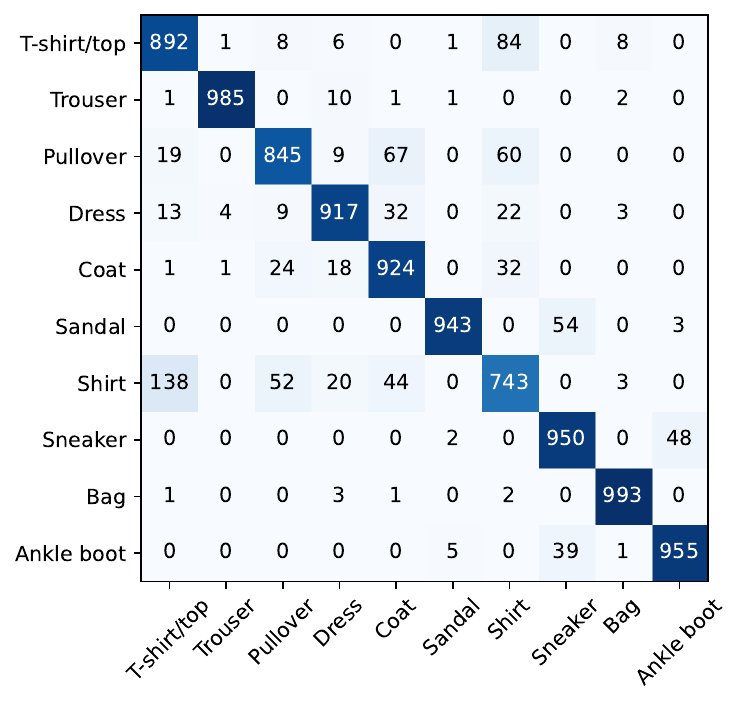}
    \caption{Without RL}
    \label{fig:CM-fmnist-woRL.}
  \end{subfigure}  
  \begin{subfigure}[b]{0.49\linewidth}
    \includegraphics[width=\textwidth]{imgs/cm/cm_fmnist_DisCon.pdf}
    \caption{CAD}
    \label{fig:CM-DisCon-ab}
  \end{subfigure}
  \caption{Ablation of Confusion Matrix on the Fashion-MNIST benchmark.}
  \label{fig:CM-ablation}
\end{figure}

\paragraph{Confusion matrix for Fashion-MNIST.}As a complementary for confusion matrix experiments in the main paper, in Figure \ref{fig:CM-fmnist}, we further provide the confusion matrix for Fashion-MNIST benchmarks. And in Figure \ref{fig:CM-ablation}, we provide the ablation results for confusion matrix in Fashion-MNIST, results shows that both stretegis we used are contributed for class confusion.

\begin{figure}[ht]
  \centering
  \begin{subfigure}[b]{0.49\linewidth}
    \includegraphics[width=\textwidth]{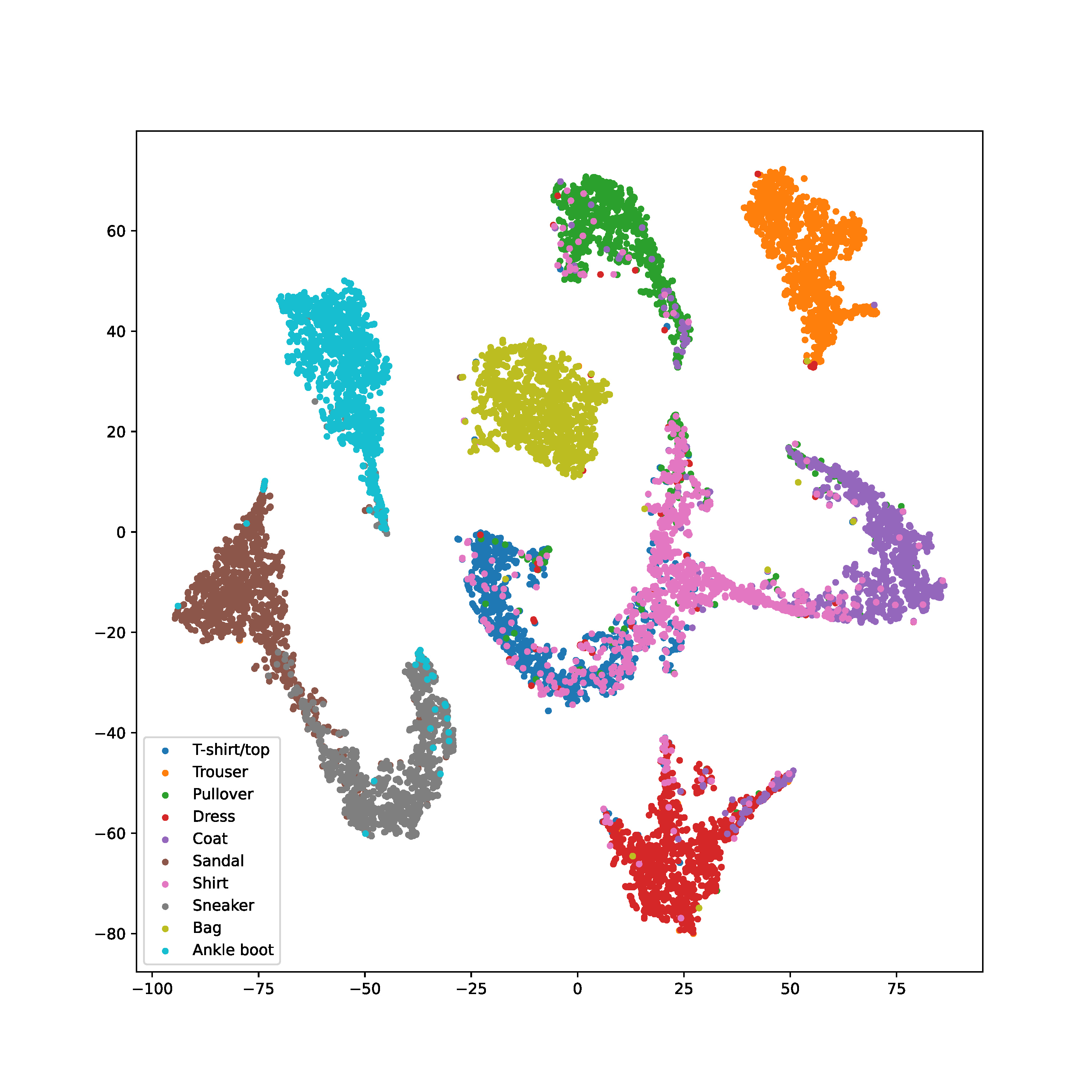}
    \caption{T-SNE of RC}
    \label{fig:TSNE-RC-fmnist}
  \end{subfigure}
  \begin{subfigure}[b]{0.49\linewidth}
    \includegraphics[width=\textwidth]{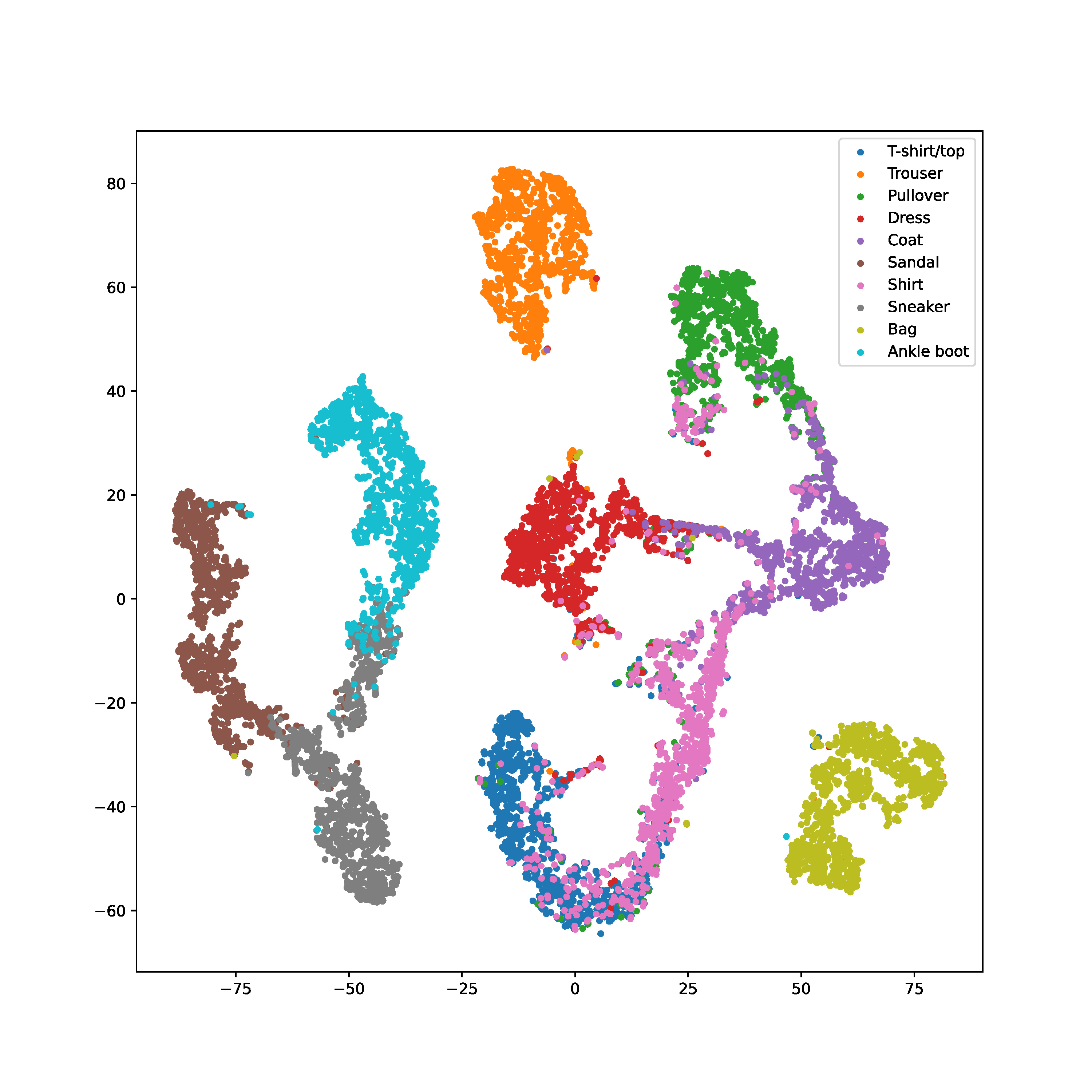}
    \caption{T-SNE of ABLE}
    \label{fig:TSNE-ABLE-fmnist}
  \end{subfigure}
  \begin{subfigure}[b]{0.49\linewidth}
    \includegraphics[width=\textwidth]{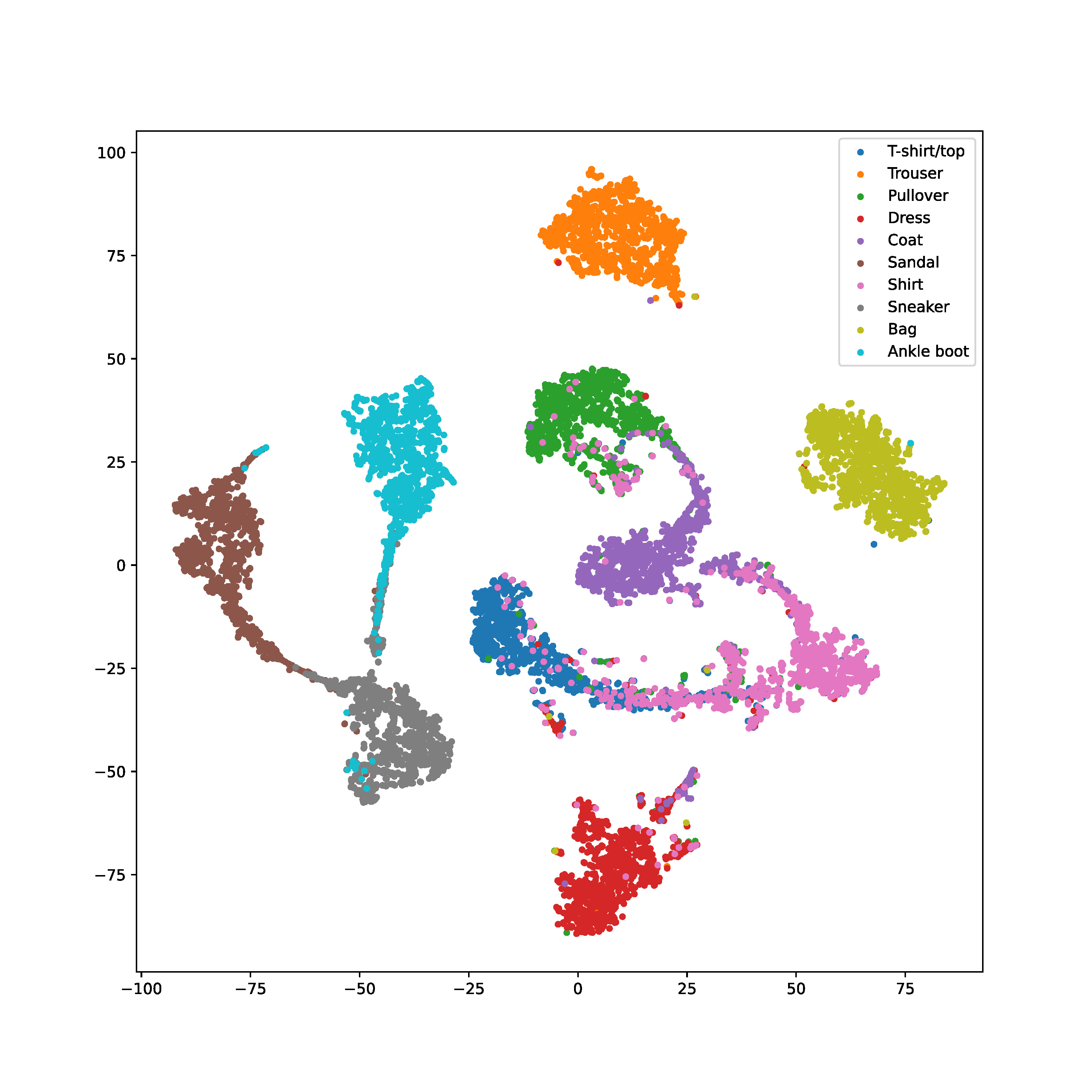}
    \caption{T-SNE of DIRK}
    \label{fig:TSNE-DIRK-fmnist}
  \end{subfigure}  
  \begin{subfigure}[b]{0.49\linewidth}
    \includegraphics[width=\textwidth]{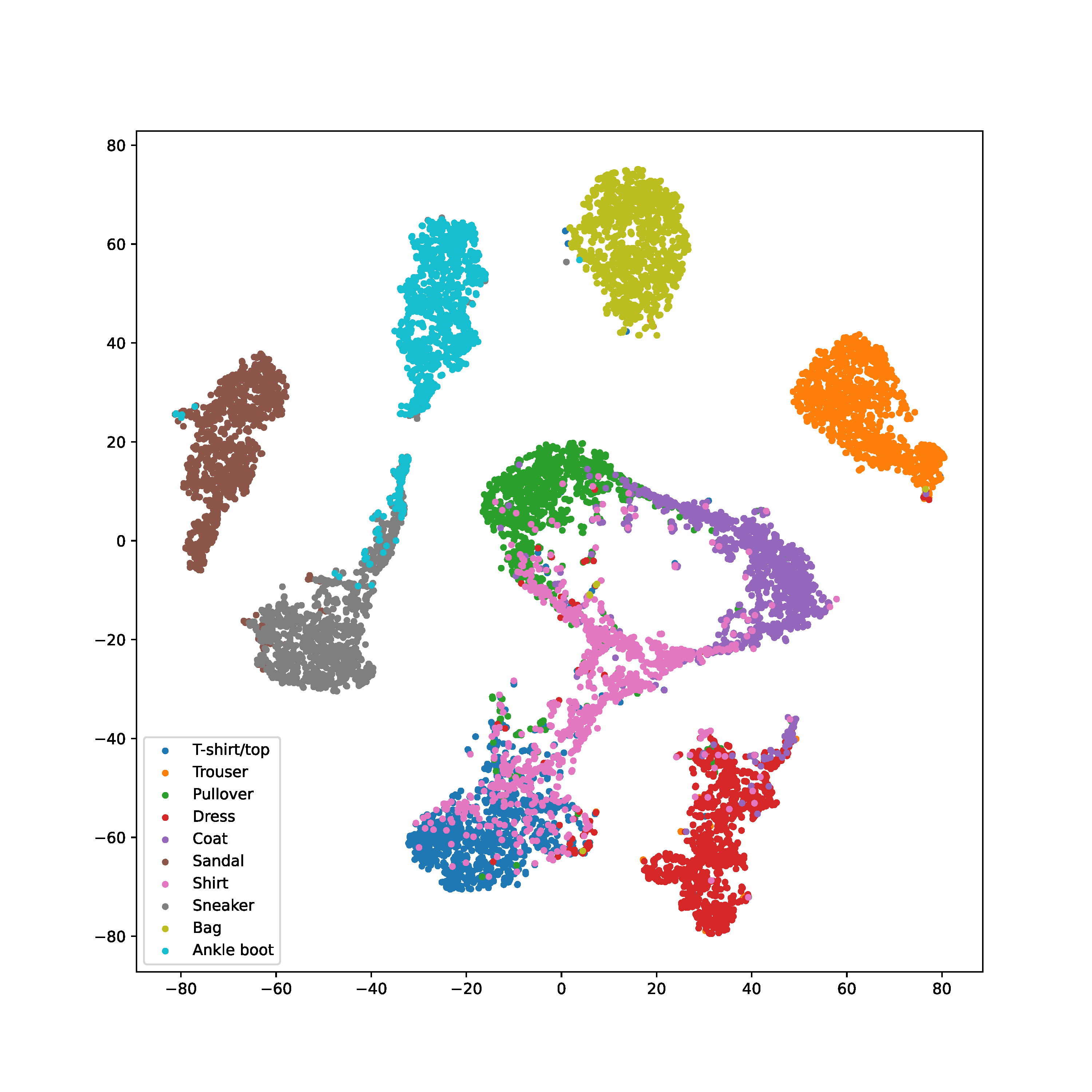}
    \caption{T-SNE of CAD}
    \label{fig:TSNE-DisCon-fmnist}
  \end{subfigure}
  \caption{T-SNE visualization on the Fashion-MNIST benchmark, with different colors representing different classes.}
  \label{fig:T-SNE-fmnist}
\end{figure}

\begin{figure}[ht]
  \centering
  \begin{subfigure}[b]{0.49\linewidth}
    \includegraphics[width=\textwidth]{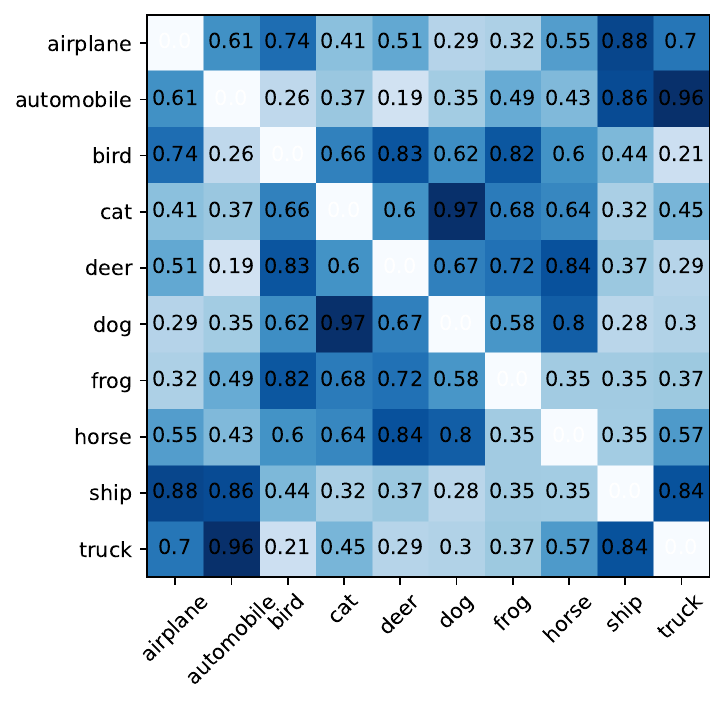}
    \caption{CIFAR-10}
    \label{fig:ol-fmnist-woBoth}
  \end{subfigure}
  \begin{subfigure}[b]{0.49\linewidth}
    \includegraphics[width=\textwidth]{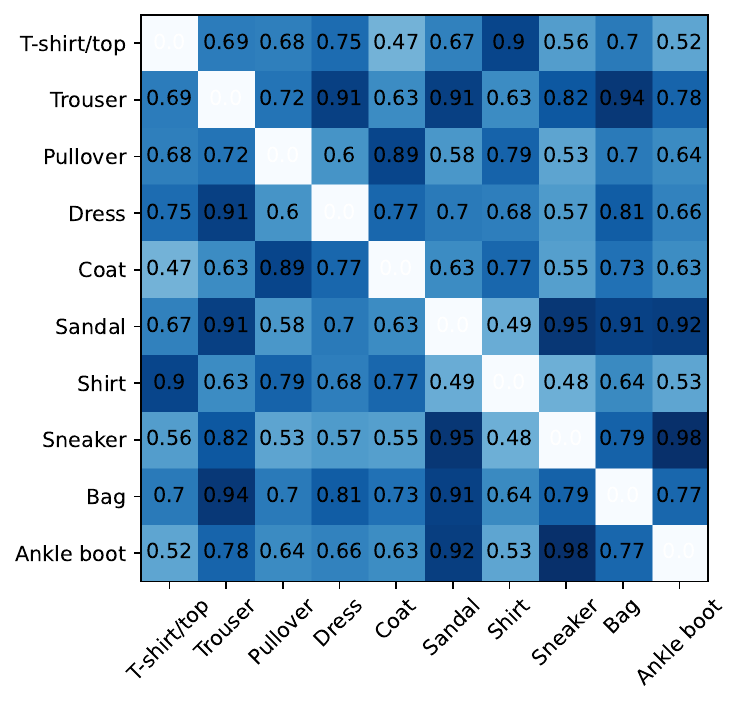}
    \caption{Fashion-MNIST}
    \label{fig:ol-fmnist-woLD}
  \end{subfigure}
  \caption{Label Overlap for the CIFAR-10 and Fashion-MNIST benchmarks.}
  \label{fig:ol-ablation}
\end{figure}

\paragraph{Label overlap for CIFAR-10 and Fashion-MNIST.} To better observe which classes are more prone to label confusion, Figure \ref{fig:ol-ablation} shows the label overlap matrices for CIFAR-10 and Fashion-MNIST, which are datasets with fewer classes for easier visualization. The element in the $i$-th row and $j$-th column of the matrix represents the proportion of instances in classes $i$ and $j$ that share both the $i$ and $j$ labels. Combined with the experimental results from the confusion matrix, we can see that most of the classes with higher label overlap are also the ones that are harder to distinguish in the confusion matrix, and our method shows significant improvement primarily on these classes.

\subsection{Parameter Sensitivity Analysis} \label{asec:parameter}
The main hyper-parameter introduced in this method is the weight $\beta$. In the main experiments, we naively set it to $1$. To analyze its impact, we provide the results with different $\beta$ on the CIFAR-10 dataset in Table \ref{tab:param}. The results show that the method is generally robust to parameter changes, with the best performance observed when $\beta$ is set in the range of $[1.00, 1.75]$. This suggests that balancing representation learning and label disambiguation in the ID-PLL task helps improve model performance.

\begin{table*}[ht]
    \centering
    \caption{Accuracy with different $\beta$ in the CIFAR-10 benchmark. The best results are highlighted in \textbf{bold}.}
    \label{tab:param}
    \setlength{\tabcolsep}{8pt}
    \begin{tabular}{cccccccccc}
        \toprule
        $\beta$  & 0.00 & 0.25 & 0.50 & 0.75 & 1.00 & 1.25 & 1.50 & 1.75 & 2.00  \\ 
        \midrule
         Accuracy & 89.19 & 92.53 & 92.95 & 92.94 & 93.85 & \textbf{94.10} & 93.97 & 93.88 & 92.87\\
        \bottomrule
    \end{tabular}
\end{table*}

Additionally, we conducted a sensitivity analysis on CIFAR-10 to assess three key contrastive learning hyperparameters:

\begin{itemize}
  \item Contrastive weight $\beta \in \{0.0, 0.5, 1.0 \text{(default)}, \textbf{1.5}, 2.0\}$ , with accuracy: 91.21\%, 93.20\%, 93.57\%, \textbf{94.10\%}, 93.64\%;
  \item Temperature $\tau \in $ \{0.03, 0.05, 0.07 \text{(default)}, 0.09, \textbf{0.12}, 0.15, 0.18\}, with accuracy: 92.64\%, 93.48\%, 93.57\%, 93.70\%, \textbf{94.23\%}, 93.71\%, 93.68\%;
  \item Momentum temperature $\tau_2 \in \{\textbf{0.1}, 0.4\text{(default)}, 0.7\}$, with accuracy: \textbf{94.58\%}, 93.57\%, 93.51\%
\end{itemize}

The default hyperparameters were inherited from prior works. Results show CAD is robust to a wide range of settings, and moderate tuning (e.g., $\beta = 1.5$, $\tau = 0.12$, $\tau_2 = 0.1$) can further improve performance.

\subsection{Experiments on Low-entanglement Conditions} \label{asec:exp_low_entangle}
To study CAD under low-entanglement conditions, we removed the annotator's most confusing class during candidate construction. In this case, CAD achieves 95.60\% on CIFAR-10, comparable to DIRK (95.65\%) and ABLE (95.31\%). These results support the paper's claim that instance entanglement is a key challenge in partial-label learning. When entanglement is reduced, label ambiguity decreases, and all methods tend to converge-yet CAD remains competitive, confirming its robustness across varying entanglement levels.

\subsection{Experiments on text classification} \label{asec:exp_text}
To evaluate the generalization of CAD beyond vision, we applied CAD to the AGNews text classification task using BERT-base as the backbone encoder. For class-specific data augmentation, we employed T5-base \cite{raffel2020exploring} to generate synthetic samples via prompt-based rewriting. Specifically, for each original news article labeled with class $c$ (e.g., “World”, “Sports”, “Business”, or “Sci/Tech”), we used one of the following class-conditional prompts:

\begin{itemize}
  \item Rewrite this article with richer $\{\text{class}\}$ terms:
  \item Restyle the following report using $\{\text{class}\}$ language:
  \item Retell the following article using $\{\text{class}\}$ phrases:
\end{itemize}

where $\{\text{class}\}$ is replaced by the name of the corresponding augmented class. The generated texts are then used as augmented training samples within CAD's feature alignment framework. As shown in Table~\ref{tab:text_results}, CAD outperforms both ABLE and DIRK, demonstrating its generalization capability to non-vision domains such as text classification.

\begin{table}[ht]
  \centering
  \caption{Text classification accuracy (\%) on AGNews. All methods use BERT-base as the student encoder; augmented data are generated using T5-base with class-specific prompts.}
  \label{tab:text_results}
  \begin{tabular}{lcc}
  \toprule
  Method & AGNews Accuracy (\%) \\
  \midrule
  DIRK   & 77.34 \\
  ABLE   & 93.92 \\
  CAD (Ours) & \textbf{94.30} \\
  \bottomrule
\end{tabular}
\end{table}

\subsection{Recovered Rate under Full Supervision} \label{asec:fraction}

We report the recovery rate, i.e., the fraction of entangled instances misclassified but correctly classified under full supervision. Table~\ref{tab:recovered_rate} shows the recovery rate for a representative baseline (DIRK), showing values up to 79.43\% (CIFAR-100, $\xi=0.9$), indicating these samples are learnable under full supervision but hindered by entanglement.

\begin{table}[htb]
    \vspace{-2pt}
    \centering
    \caption{Recovery rate (\%) of entangled instances misclassified by DIRK but correctly classified under full supervision (ground-truth labels), at different similarity thresholds $\xi$.}
    \label{tab:recovered_rate}
    \setlength{\tabcolsep}{6pt}
    \renewcommand{\arraystretch}{1.15}
    \small
    \begin{tabular}{lccc}
        \toprule
        $\xi$ & F-MNIST & CIFAR-10 & CIFAR-100 \\
        \midrule
        0.90 & 16.01 & 27.45 & 79.43 \\
        0.95 & 15.28 & 26.56 & 51.06 \\
        0.99 & 11.51 & 25.00 & 38.46 \\
        \bottomrule
    \end{tabular}
    \vspace{-6pt}
\end{table}

\section{Visualization Analysis} 
\subsection{Augmentation Analysis} 
\label{asec:visual}
This section presents a visualization analysis of augmentations produced by diffusion-based and CAM-based methods, with the aim of assessing whether each method effectively emphasizes class-discriminative regions.

\begin{figure}[tb]
  \centering
  \includegraphics[width=0.99\linewidth]{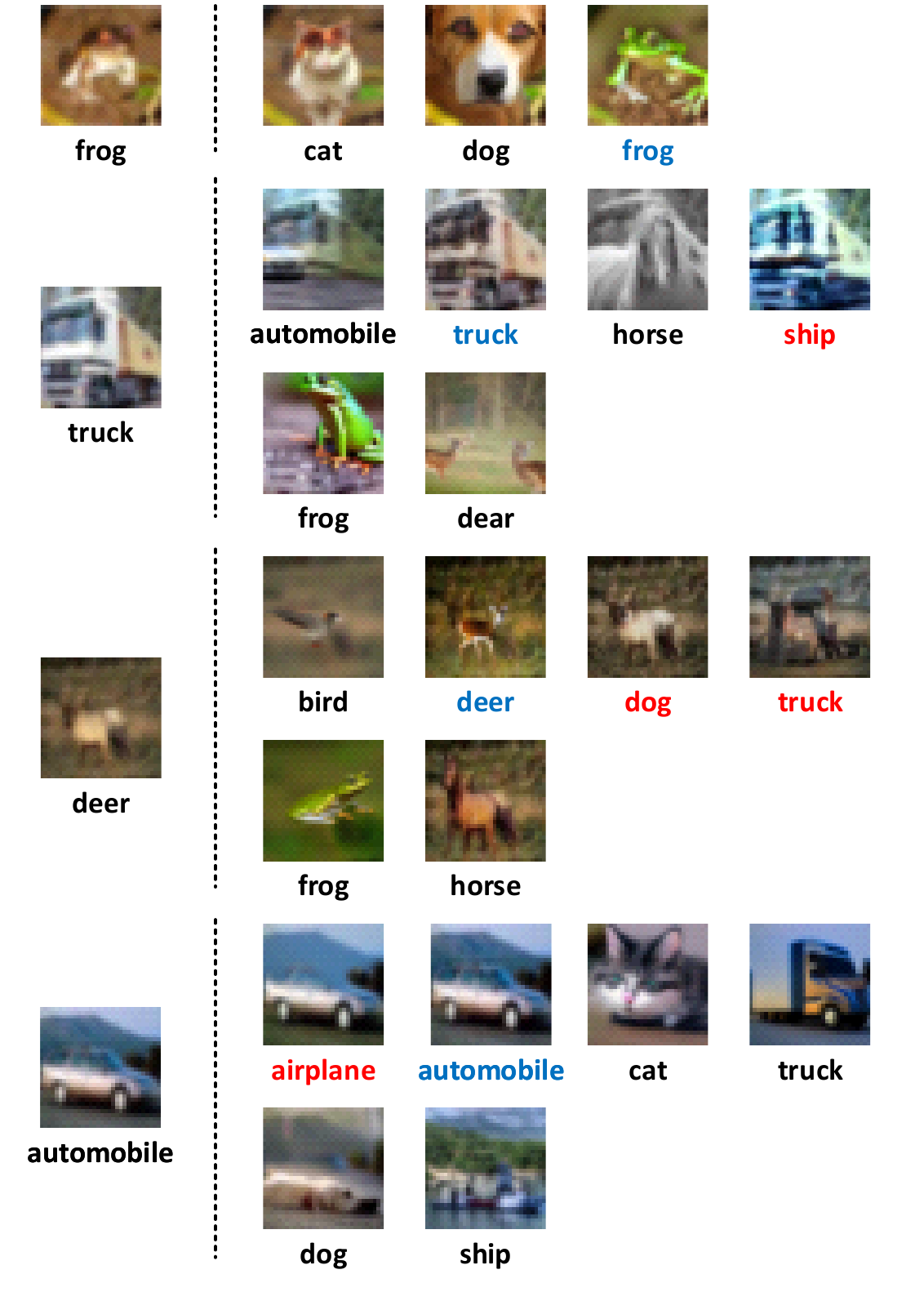} 
  \caption{Diffusion-based visualization of the class-specific augmentations generated from CIFAR-10. The left side shows the samples with their ground truth labels, while the right side displays the augmentations generated with different candidate label guidance.}
  \label{fig:visualization}
\end{figure}

\paragraph{Analysis for diffusion-based augmentation.} To intuitively analyze whether the generated augmentations are helpful, we present some of the visualization results in CIFAR-10 benchmarks. Figure \ref{fig:visualization} shows four sets of examples: a frog instance labeled as \{cat, dog, frog\}, a truck instance labeled as \{truck, automobile, horse, frog, deer, ship\}, a deer instance labeled as \{bird, deer, dog, frog, horse, truck\}, and an automobile instance labeled as \{airplane, automobile, cat, dog, ship, truck\}. It can be observed that the augmentations generated by guiding instances with their ground-truth labels (marked in blue) generally do not show significant changes. This helps preserve the original information of the instance in contrastive learning. For other class-specific augmentations, we can see that most of them exhibit some characteristics of the corresponding class while preserving the original instance's color and texture. Maximizing the preservation of instance-specific characteristics while enhancing classes-specific characteristics can help the model better distinguish between class-related features and the instance's unique traits. 

However, some of these augmentations exhibit noticeable defects (marked in red): while class-related features are successfully amplified—such as the oceanic blue tones and reflective gloss in the ship-like example (second row)—the edits often introduce artifacts or semantic inconsistencies. Directly adding such diffusion-edited instances to existing methods and treating the induced labels as supervision degrades performance, largely due to these imperfections and the resulting distribution shift. This indicates that simply injecting edited samples is insufficient; instead, their utility depends on how they are structurally integrated into the learning objective. Therefore, rather than using these imperfect augmentations directly for classifier training, we employ them in representation learning with quality-based reweighting, which allows us to harness their amplified class signals while mitigating the negative impact of their imperfections.

Furthermore, to quantify semantic alignment, we have computed the CLIP-score margin $m(\boldsymbol{x}') = \text{sim}(\boldsymbol{x}', t) - \text{sim}(\boldsymbol{x}', y)$ for a generated sample $\boldsymbol{x}'$, comparing its similarity to the target class $t$ versus the original class $y$. As shown in Table \ref{tab:clipscore}, margins flip from negative to positive with a +17.21\% avg. in target CLIP-score, confirming improved alignment with the target semantics.

\begin{table*}[ht]
    \vspace{-8pt}
    \centering
    \caption{Comparison of CLIP-score margin and target similarity (margin $\mid$ CLIP-score). } \label{tab:clipscore}
    \setlength{\tabcolsep}{3pt}
    \begin{tabular}{lccccc}
        \toprule
        Method & Fashion-MNIST & CIFAR-10 & CIFAR-100 & Flower & Oxford-IIIT Pet \\
        \midrule
        Original ($x$) & -0.0276 $\mid$ 0.2364 & -0.0415 $\mid$ 0.2151 & -0.0416 $\mid$ 0.2176 & -0.0010 $\mid$ 0.2185 & -0.0002 $\mid$ 0.2282 \\
        Diffusion ($x'$) & \textbf{0.0120 $\mid$ 0.2555} & \textbf{0.0314 $\mid$ 0.2566} & \textbf{0.0181 $\mid$ 0.2453} & \textbf{0.0243 $\mid$ 0.2784} & \textbf{0.0169 $\mid$ 0.2705} \\
        \bottomrule
    \end{tabular}
    \vspace{-10pt}
\end{table*}

\begin{figure}[tb]
  \centering
  \includegraphics[width=0.99\linewidth]{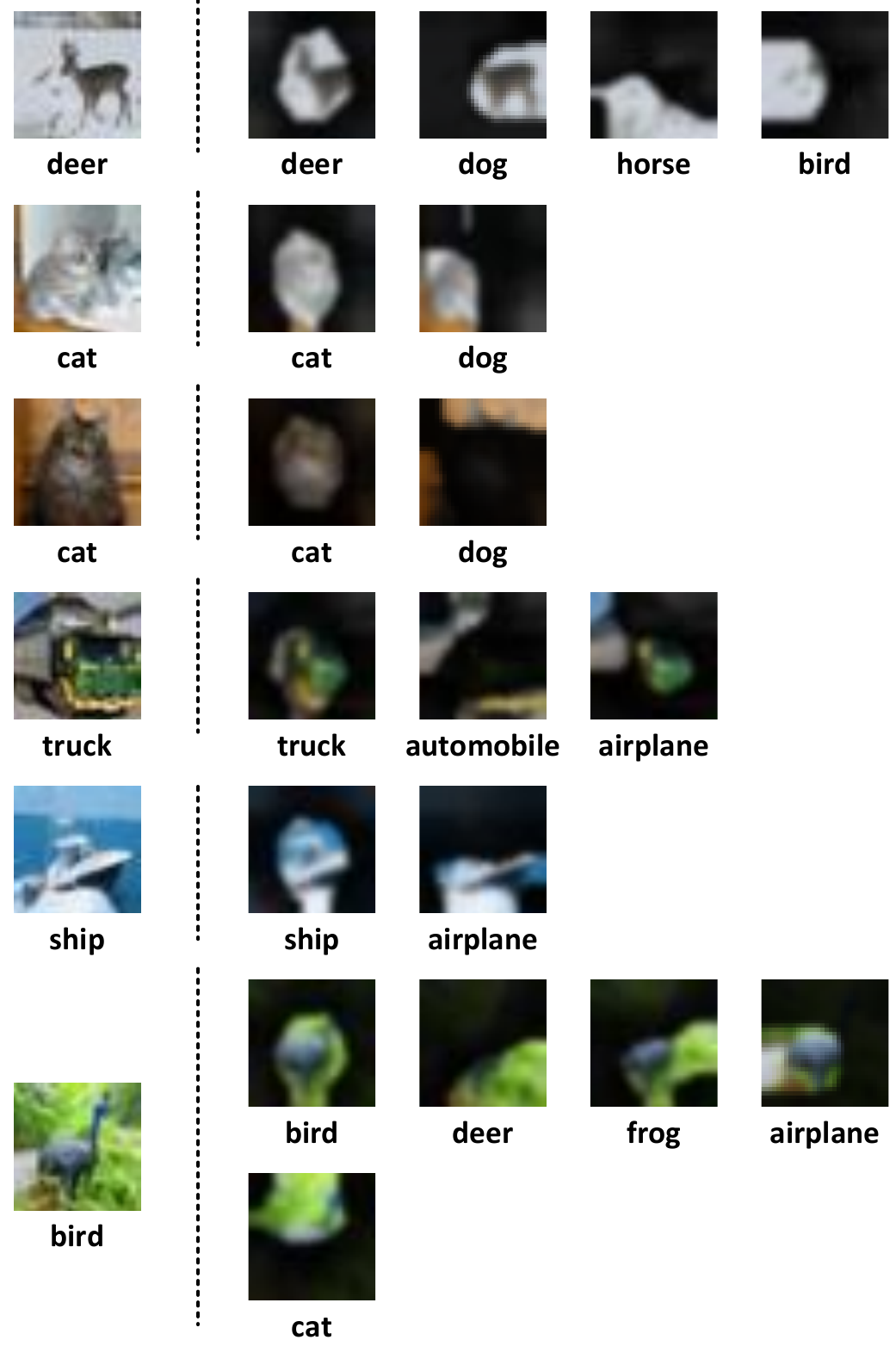} 
  \caption{CAM-based visualization of the class-specific augmentations generated from CIFAR-10. The left side shows the samples with their ground truth labels, while the right side displays the augmentations generated with different candidate label guidance.}
  \label{fig:visualization_cam}
\end{figure}

\paragraph{Analysis for CAM-based augmentation.} 
Figure~\ref{fig:visualization_cam} presents several examples of CAM-based augmentations generated from CIFAR-10. Here, we show the augmented samples actually used during training—specifically, those for which a salient activation region exists (as described in Appendix~\ref{asec:cadcam}, samples with nearly uniform activation maps are discarded).

Despite lacking explicit semantic editing capabilities, the CAM-based strategy effectively amplifies class-relevant features by suppressing non-discriminative regions. For instance, in the first row, when the guiding label is deer, the augmentation preserves the deer's head and body; when guided by the label dog, only parts resembling a dog's torso and limbs are retained. Similarly, for the truck example in the fourth row, using truck as the guide retains most of the cab, whereas the automobile label leads to preservation of only the lower part of the vehicle along with some metallic background structures. When guided by airplane, the augmentation highlights a side panel of the trailer whose shape resembles an aircraft wing. 

After obtaining these class-specific augmentations, we align only augmentation pairs that share the same guiding label. Because each view retains solely the regions activated by that label, the resulting contrastive alignment operates between semantically consistent local features, thereby avoiding the cross-class mismatches typical of direct instance-level alignment.

\subsection{T-SNE for Fashion-MNIST} 
\label{asec:tsne}
In this section, we provide the T-SNE visualization for Fashion-MNIST as a complement to the experiments in Figure \ref{fig:t-SNE} of the main paper. The results show that our proposed method has clearer class boundaries. A notable example is the clusters of blue, brown, and gray colors on the left. In CAD (Figure \ref{fig:TSNE-DisCon-fmnist}), although there are some errors, the boundaries of the three clusters are well-separated. In contrast, while DIRK (Figure \ref{fig:TSNE-DIRK-fmnist}) achieves similar accuracy to CAD, the boundary samples of the three clusters are mixed together. This demonstrates that CAD effectively encourages clearer class boundaries. Overall, RC exhibits the most color overlap, which is due to its significantly lower accuracy compared to the other three methods. While the color overlap in ABLE and DIRK is less than that in RC, the confusion of boundary samples has not improved. In fact, on Fashion-MNIST, it is even worse than RC. This is because both ABLE and DIRK use candidate labels to guide contrastive learning, leading to negative effects from the misalignment of samples near similar class boundaries. In contrast, CAD shows the clearest boundaries on both benchmarks, demonstrating that the proposed strategy effectively manages class boundaries and reduces class confusion.

\section{Additional Related Work}
\label{asec:relateddiff}

\subsection{Partial Label Learning}
Partial Label Learning, also known as Ambiguous Label learning \cite{hullermeier2006learning,chen2017learning,li2021detecting} or Superset Label learning \cite{liu2014learnability,liu2012conditional}, is a weakly supervised learning task where each training instance is annotated with a set of candidate labels that includes the ground truth. Early PLL research used a straightforward average-based strategy that treated each candidate label equally \cite{hullermeier2006learning,cour2011learning}. However, the ambiguity in the labels could mislead the model's training. To address this issue, identification-based strategies, which explicitly handle ambiguous labels, have received attention \cite{jin2002learning,liu2012conditional,yu2016maximum}. These studies have introduced numerous learning techniques to aid in the disambiguation, such as reweighting \cite{tang2017confidence,feng2020provably,wen2021leveraged}, employing k-nearest neighbors \cite{hullermeier2006learning,zhang2015solving}, maximizing margins \cite{yu2016maximum,nguyen2008classification}, self-training \cite{tian2024crosel}, graph learning \cite{lyu2022deep,wang2019adaptive}, meta-learning \cite{xie2021partial}, and contrastive learning \cite{wang2022pico}. Most of these methods assume that candidate labels are sampled independently of the instance, which hinders their application in real-world scenarios.

Recently, many studies have explored instance-dependent PLL tasks \cite{xu2021instance,xia2022ambiguity,qiao2022decompositional,xu2023progressive,he2023candidate,wu2024distilling}. For example, VALEN \cite{xu2021instance} restores the underlying label distribution by inferring the posterior density using Dirichlet density. ABLE \cite{xia2022ambiguity} leverages instance-dependent characteristics to assist label disambiguation through ambiguity-guided contrastive learning. DIRK \cite{wu2024distilling} constructs reliable pseudo-labels by controlling the confidence of negative labels. However, we note that these works lack explicit governance of similar class confusion, which is a common problem in ID-PLL. It should be noted that although the early maximum margin strategies can implicitly control the distance between similar classes \cite{yu2016maximum,nguyen2008classification}, these methods face challenges in scalability to large-scale high-dimensional data, while lacking entanglement handling from both representation learning and confidence adjustment. 

\subsection{Contrastive Learning}
Contrastive learning is a representation learning technique that works by bringing similar instances closer together and pushing dissimilar instances farther apart in the representation space. It has demonstrated its effectiveness across a wide range of supervised and unsupervised representation learning tasks \cite{oord2018representation,chen2020simple,he2020momentum,chen2020improved,khosla2020supervised}. Building on this success, contrastive learning has naturally been applied to many weakly supervised learning tasks, such as complementary label learning \cite{ruan2022biased}, noisy label learning \cite{li2022selective,ghosh2021contrastive}, and semi-supervised learning \cite{sohn2020fixmatch}. A crucial aspect of these methods is identifying pairs of instances from the same class (referred to as positive pairs) and guiding them towards aligned representations, which integrates contrastive learning with supervised information. In recent years, some methods \cite{wang2022pico,xia2022ambiguity,wu2024distilling} have explored adapting supervised contrastive learning to PLL. These methods often rely on pseudo-labels or feature distances based on model outputs to find the positive pairs. However, in ID-PLL, instances from similar classes often share the same labels, making them prone to mistaking them as false positive pairs. In this paper, unlike previous approaches, we treat class-specific augmentations bootstrapped by the same label as positive pairs to mitigate such false positives.

\subsection{Diffusion-based Image Editing}
Diffusion models, inspired by principles of non-equilibrium thermodynamics, generate data matching the original distribution by progressively adding noise to the data and learning the reverse process \cite{huang2024diffusion}. In recent years, diffusion models have gained particular interest in image editing due to their greater control compared to the GAN-based methods. Depending on the type of editing, these studies can be categorized into semantic editing \cite{kim2022diffusionclip,kwon2022diffusion,xu2024cyclenet,preechakul2022diffusion,zhao2022egsde,brooks2023instructpix2pix,guo2024focus,geng2024instructdiffusion,zhang2024hive,huberman2024edit,brack2024ledits++,pan2023effective}, stylistic editing \cite{kim2022diffusionclip,tumanyan2023plug,wang2023stylediffusion,brooks2023instructpix2pix,wu2023uncovering,geng2024instructdiffusion,bodur2024iedit,huberman2024edit,wallace2023edict,hertz2022prompt}, and structural editing \cite{li2023layerdiffusion,epstein2023diffusion,patashnik2023localizing,epstein2023diffusion}. Instructional editing \cite{brooks2023instructpix2pix,li2023moecontroller,guo2024focus,zhang2024magicbrush,geng2024instructdiffusion} is a key direction in the semantic editing domain, enabling users to guide the image editing process through natural language instructions. For example, InstructPix2Pix \cite{brooks2023instructpix2pix} first achieves image content editing by constructing an instruction-image paired dataset. Building on this, MoEController \cite{li2023moecontroller} and FOI \cite{guo2024focus} enhance editing accuracy and efficiency by introducing expert modules and improving implicit alignment capabilities, respectively. Recently, methods like MagicBrush \cite{zhang2024magicbrush} and InstructDiffusion \cite{geng2024instructdiffusion} have further improved image editing performance and diversity through more comprehensive dataset construction and enhancement strategies. MagicBrush enriches its dataset by employing human editing sessions, while InstructDiffusion expands its data using tool-generated content and Photoshop requests sourced from the internet. Notably, the concept of diffusion has recently gained attention in some weakly- or semi-supervised tasks \cite{chen2024label,ho2023diffusion,you2024diffusion}.

\section{Prompt with Cues for Oxford-IIIT Pet} \label{asec:promptcues}
\begin{itemize}
  \item Abyssinian: sleek medium-sized cat with short dense ticked reddish-brown coat, almond-shaped green or gold eyes, large pointed ears, slender muscular body, no stripes or spots.
  \item American Bulldog: large muscular dog with broad head, strong jaws, short white or white-with-patches coat, loose facial skin, powerful stance.
  \item American Pit Bull Terrier: medium-sized athletic dog with short glossy coat in various colors, broad head, strong neck, defined cheek muscles, short tail, alert agile build.
  \item Basset Hound: low-slung heavy-boned dog with extremely long ears, droopy eyes, loose skin, short legs, tricolor (black-white-tan) coat, long curved tail.
  \item Beagle: small to medium hound with short tricolor (black-white-tan) coat, large floppy ears, expressive brown eyes, compact muscular body, white-tipped upright tail.
  \item Bengal: medium to large cat with short sleek spotted or marbled golden/brown/silver coat, muscular build, green or gold eyes, wild leopard-like pattern.
  \item Birman: medium-sized cat with semi-long silky white or cream coat, colorpoint markings, deep blue eyes, distinctive white 'gloves' on all four paws.
  \item Bombay: medium-sized cat with short jet-black glossy coat, copper or gold eyes, rounded head, sleek muscular body, panther-like appearance.
  \item Boxer: medium-large square-built dog with short fawn or brindle coat, black facial mask, undershot jaw, docked tail, alert energetic expression.
  \item British Shorthair: stocky round cat with dense short blue-gray or solid-color coat, large round copper eyes, broad face with full cheeks, plush teddy-bear look.
  \item Chihuahua: tiny dog with large erect ears, apple-shaped head, smooth or long coat in various colors, prominent eyes, often tail curled over back.
  \item Egyptian Mau: medium cat with short spotted silver/bronze/smoke coat, green gooseberry eyes, 'M'-shaped forehead marking, athletic graceful build.
  \item English Cocker Spaniel: medium sporting dog with long silky feathered coat (parti-color or solid), long pendulous ears, soft expressive eyes, docked tail.
  \item English Setter: elegant medium-large gundog with long feathered white coat speckled in black/orange/lemon ('belton'), long ears, gentle refined face.
  \item German Shorthaired Pointer: lean athletic hunting dog with short liver-and-white or solid liver coat, docked tail, long muzzle, muscular build, alert intelligent look.
  \item Great Pyrenees: very large majestic dog with thick pure white double coat, dark eyes, often double dewclaws on hind legs, calm bear-like appearance.
  \item Havanese: small toy dog with long silky wavy or curly coat in various colors, plumed tail over back, dark eyes, lively cheerful expression.
  \item Japanese Chin: tiny flat-faced toy dog with long silky black-and-white or red-and-white coat, large wide-set eyes, short muzzle, feathered arched tail.
  \item Keeshond: medium spitz with thick gray-black-cream double coat, facial 'spectacles' markings, curled tail, foxy expression, abundant neck ruff.
  \item Leonberger: giant dog with lion-like mane (males), long water-resistant golden-yellow to red-brown coat, black mask, webbed feet, gentle giant look.
  \item Maine Coon: large rugged cat with long shaggy coat, tufted ears and paws, bushy tail, rectangular body, lynx-like ear tips, friendly expression.
  \item Miniature Pinscher: small compact dog with short smooth red/black-and-tan/chocolate coat, erect ears, docked tail, high-stepping gait, alert fearless look.
  \item Newfoundland: massive heavy-boned dog with thick water-resistant black/brown/white-and-black double coat, webbed feet, droopy jowls, sweet calm expression.
  \item Persian: round flat-faced cat with extremely long thick coat, small ears, large round copper eyes, cobby body, doll-like face.
  \item Pomeranian: tiny spitz with abundant fluffy double coat (often orange), fox-like face, small erect ears, plumed tail curled over back.
  \item Pug: small square dog with wrinkled face, short muzzle, curled tail, smooth fawn or black coat, large dark eyes, compact muscular body.
  \item Ragdoll: large semi-longhaired cat with soft plush coat, blue eyes, colorpoint pattern (seal/blue/etc.), floppy when held, gentle expression.
  \item Russian Blue: medium cat with short dense silvery-blue double coat, green eyes, wedge-shaped head, elegant fine-boned build, graceful posture.
  \item Saint Bernard: enormous dog with thick red-and-white or mahogany-and-white coat, massive head with droopy jowls, calm benevolent expression, short or long coat variant.
  \item Samoyed: medium-large spitz with thick pure white double coat, black lips/nose, 'smiling' expression, curled tail, erect triangular ears.
  \item Scottish Terrier: small sturdy terrier with wiry black (or brindle/wheaten) coat, long beard and eyebrows, short legs, dignified independent look.
  \item Shiba Inu: small fox-like dog with short red/sesame/black-and-tan coat, curled tail, triangular erect ears, bold alert expression, compact agile build.
  \item Siamese: sleek slender cat with short cream coat and dark colorpoint ears/face/paws/tail, almond-shaped blue eyes, wedge-shaped head, elegant posture.
  \item Sphynx: hairless cat with wrinkled skin, large lemon-shaped eyes, prominent cheekbones, large ears, muscular body, may have fine downy fuzz.
  \item Staffordshire Bull Terrier: compact muscular dog with short smooth coat in various colors, broad head with pronounced cheek muscles, dark eyes, short tail, affectionate look.
  \item Wheaten Terrier: medium terrier with soft wavy wheaten (golden-beige) coat, dark eyes, square head, topknot of longer hair, cheerful expression.
  \item Yorkshire Terrier: tiny toy dog with long silky steel-blue and tan coat, small V-shaped ears, compact body, tail held high, elegant feisty look."
\end{itemize}

%% file: main.bib
@String(IJCAI = {IJCAI})

@String(AAAI = {AAAI})

@article{jin2002learning,
  title   = {Learning with multiple labels},
  author  = {Jin, Rong and Ghahramani, Zoubin},
  journal = {Advances in neural information processing systems},
  volume  = {15},
  year    = {2002}
}

@article{hullermeier2006learning,
  title     = {Learning from ambiguously labeled examples},
  author    = {H{\"u}llermeier, Eyke and Beringer, J{\"u}rgen},
  journal   = {Intelligent Data Analysis},
  volume    = {10},
  number    = {5},
  pages     = {419--439},
  year      = {2006},
  publisher = {IOS Press}
}

@inproceedings{nguyen2008classification,
  title     = {Classification with partial labels},
  author    = {Nguyen, Nam and Caruana, Rich},
  booktitle = {Proceedings of the 14th ACM SIGKDD international conference on Knowledge discovery and data mining},
  pages     = {551--559},
  year      = {2008}
}

@inproceedings{nilsback2008automated,
  title        = {Automated flower classification over a large number of classes},
  author       = {Nilsback, Maria-Elena and Zisserman, Andrew},
  booktitle    = {2008 Sixth Indian conference on computer vision, graphics \& image processing},
  pages        = {722--729},
  year         = {2008},
  organization = {IEEE}
}

@article{van2008visualizing,
  title   = {Visualizing data using t-SNE.},
  author  = {Van der Maaten, Laurens and Hinton, Geoffrey},
  journal = {Journal of machine learning research},
  volume  = {9},
  number  = {11},
  year    = {2008}
}

@article{krizhevsky2009learning,
  title   = {Learning multiple layers of features from tiny images},
  author  = {Krizhevsky, Alex and Hinton, Geoffrey and others},
  year    = {2009},
  journal = {Master's thesis, University of Tront}
}

@article{cour2011learning,
  title     = {Learning from partial labels},
  author    = {Cour, Timothee and Sapp, Ben and Taskar, Ben},
  journal   = {The Journal of Machine Learning Research},
  volume    = {12},
  pages     = {1501--1536},
  year      = {2011},
  publisher = {JMLR. org}
}

@inproceedings{briggs2012rank,
  title     = {Rank-loss support instance machines for MIML instance annotation},
  author    = {Briggs, Forrest and Fern, Xiaoli Z and Raich, Raviv},
  booktitle = {Proceedings of the 18th ACM SIGKDD international conference on Knowledge discovery and data mining},
  pages     = {534--542},
  year      = {2012}
}

@inproceedings{parkhi2012cats,
  title        = {Cats and dogs},
  author       = {Parkhi, Omkar M and Vedaldi, Andrea and Zisserman, Andrew and Jawahar, CV},
  booktitle    = {2012 IEEE conference on computer vision and pattern recognition},
  pages        = {3498--3505},
  year         = {2012},
  organization = {IEEE}
}

@article{liu2012conditional,
  title   = {A conditional multinomial mixture model for superset label learning},
  author  = {Liu, Liping and Dietterich, Thomas},
  journal = {Advances in neural information processing systems},
  volume  = {25},
  year    = {2012}
}

@inproceedings{liu2014learnability,
  title        = {Learnability of the superset label learning problem},
  author       = {Liu, Liping and Dietterich, Thomas},
  booktitle    = {International conference on machine learning},
  pages        = {1629--1637},
  year         = {2014},
  organization = {PMLR}
}

@inproceedings{zhang2015solving,
  title     = {Solving the partial label learning problem: An instance-based approach.},
  author    = {Zhang, Min-Ling and Yu, Fei},
  booktitle = {IJCAI},
  pages     = {4048--4054},
  year      = {2015}
}

@inproceedings{yu2016maximum,
  title        = {Maximum margin partial label learning},
  author       = {Yu, Fei and Zhang, Min-Ling},
  booktitle    = {Asian conference on machine learning},
  pages        = {96--111},
  year         = {2016},
  organization = {PMLR}
}

@inproceedings{he2016identity,
  title        = {Identity mappings in deep residual networks},
  author       = {He, Kaiming and Zhang, Xiangyu and Ren, Shaoqing and Sun, Jian},
  booktitle    = {Computer Vision--ECCV 2016: 14th European Conference, Amsterdam, The Netherlands, October 11--14, 2016, Proceedings, Part IV 14},
  pages        = {630--645},
  year         = {2016},
  organization = {Springer}
}

@inproceedings{zhou2016learning,
  title     = {Learning deep features for discriminative localization},
  author    = {Zhou, Bolei and Khosla, Aditya and Lapedriza, Agata and Oliva, Aude and Torralba, Antonio},
  booktitle = {Proceedings of the IEEE conference on computer vision and pattern recognition},
  pages     = {2921--2929},
  year      = {2016}
}

@article{chen2017learning,
  title     = {Learning from ambiguously labeled face images},
  author    = {Chen, Ching-Hui and Patel, Vishal M and Chellappa, Rama},
  journal   = {IEEE transactions on pattern analysis and machine intelligence},
  volume    = {40},
  number    = {7},
  pages     = {1653--1667},
  year      = {2017},
  publisher = {IEEE}
}

@article{xiao2017fashion,
  title   = {Fashion-mnist: a novel image dataset for benchmarking machine learning algorithms},
  author  = {Xiao, Han and Rasul, Kashif and Vollgraf, Roland},
  journal = {arXiv preprint arXiv:1708.07747},
  year    = {2017}
}

@inproceedings{tang2017confidence,
  title     = {Confidence-rated discriminative partial label learning},
  author    = {Tang, Cai-Zhi and Zhang, Min-Ling},
  booktitle = {Proceedings of the AAAI conference on artificial intelligence},
  volume    = {31},
  year      = {2017}
}

@article{devries2017improved,
  title   = {Improved regularization of convolutional neural networks with cutout},
  author  = {DeVries, Terrance and Taylor, Graham W},
  journal = {arXiv preprint arXiv:1708.04552},
  year    = {2017}
}

@article{cubuk2018autoaugment,
  title   = {Autoaugment: Learning augmentation policies from data},
  author  = {Cubuk, Ekin D and Zoph, Barret and Mane, Dandelion and Vasudevan, Vijay and Le, Quoc V},
  journal = {arXiv preprint arXiv:1805.09501},
  year    = {2018}
}

@article{oord2018representation,
  title   = {Representation learning with contrastive predictive coding},
  author  = {Oord, Aaron van den and Li, Yazhe and Vinyals, Oriol},
  journal = {arXiv preprint arXiv:1807.03748},
  year    = {2018}
}

@inproceedings{zhang2019multiple,
  title     = {Multiple Noisy Label Distribution Propagation for Crowdsourcing.},
  author    = {Zhang, Hao and Jiang, Liangxiao and Xu, Wenqiang},
  booktitle = {IJCAI},
  pages     = {1473--1479},
  year      = {2019}
}

@inproceedings{wang2019adaptive,
  title     = {Adaptive graph guided disambiguation for partial label learning},
  author    = {Wang, Deng-Bao and Li, Li and Zhang, Min-Ling},
  booktitle = {Proceedings of the 25th ACM SIGKDD international conference on knowledge discovery \& data mining},
  pages     = {83--91},
  year      = {2019}
}

@inproceedings{cubuk2020randaugment,
  title     = {Randaugment: Practical automated data augmentation with a reduced search space},
  author    = {Cubuk, Ekin D and Zoph, Barret and Shlens, Jonathon and Le, Quoc V},
  booktitle = {Proceedings of the IEEE/CVF conference on computer vision and pattern recognition workshops},
  pages     = {702--703},
  year      = {2020}
}

@article{feng2020provably,
  title   = {Provably consistent partial-label learning},
  author  = {Feng, Lei and Lv, Jiaqi and Han, Bo and Xu, Miao and Niu, Gang and Geng, Xin and An, Bo and Sugiyama, Masashi},
  journal = {Advances in neural information processing systems},
  volume  = {33},
  pages   = {10948--10960},
  year    = {2020}
}

@inproceedings{chen2020simple,
  title        = {A simple framework for contrastive learning of visual representations},
  author       = {Chen, Ting and Kornblith, Simon and Norouzi, Mohammad and Hinton, Geoffrey},
  booktitle    = {International conference on machine learning},
  pages        = {1597--1607},
  year         = {2020},
  organization = {PMLR}
}

@article{chen2020improved,
  title   = {Improved baselines with momentum contrastive learning},
  author  = {Chen, Xinlei and Fan, Haoqi and Girshick, Ross and He, Kaiming},
  journal = {arXiv preprint arXiv:2003.04297},
  year    = {2020}
}

@inproceedings{lv2020progressive,
  title        = {Progressive identification of true labels for partial-label learning},
  author       = {Lv, Jiaqi and Xu, Miao and Feng, Lei and Niu, Gang and Geng, Xin and Sugiyama, Masashi},
  booktitle    = {international conference on machine learning},
  pages        = {6500--6510},
  year         = {2020},
  organization = {PMLR}
}

@article{khosla2020supervised,
  title   = {Supervised contrastive learning},
  author  = {Khosla, Prannay and Teterwak, Piotr and Wang, Chen and Sarna, Aaron and Tian, Yonglong and Isola, Phillip and Maschinot, Aaron and Liu, Ce and Krishnan, Dilip},
  journal = {Advances in neural information processing systems},
  volume  = {33},
  pages   = {18661--18673},
  year    = {2020}
}

@inproceedings{he2020momentum,
  title     = {Momentum contrast for unsupervised visual representation learning},
  author    = {He, Kaiming and Fan, Haoqi and Wu, Yuxin and Xie, Saining and Girshick, Ross},
  booktitle = {Proceedings of the IEEE/CVF conference on computer vision and pattern recognition},
  pages     = {9729--9738},
  year      = {2020}
}

@inproceedings{wen2021leveraged,
  title        = {Leveraged weighted loss for partial label learning},
  author       = {Wen, Hongwei and Cui, Jingyi and Hang, Hanyuan and Liu, Jiabin and Wang, Yisen and Lin, Zhouchen},
  booktitle    = {International conference on machine learning},
  pages        = {11091--11100},
  year         = {2021},
  organization = {PMLR}
}

@inproceedings{xie2021partial,
  title     = {Partial multi-label learning with meta disambiguation},
  author    = {Xie, Ming-Kun and Sun, Feng and Huang, Sheng-Jun},
  booktitle = {Proceedings of the 27th ACM SIGKDD conference on knowledge discovery \& data mining},
  pages     = {1904--1912},
  year      = {2021}
}

@article{sohn2020fixmatch,
  title   = {Fixmatch: Simplifying semi-supervised learning with consistency and confidence},
  author  = {Sohn, Kihyuk and Berthelot, David and Carlini, Nicholas and Zhang, Zizhao and Zhang, Han and Raffel, Colin A and Cubuk, Ekin Dogus and Kurakin, Alexey and Li, Chun-Liang},
  journal = {Advances in neural information processing systems},
  volume  = {33},
  pages   = {596--608},
  year    = {2020}
}

@article{raffel2020exploring,
  title   = {Exploring the limits of transfer learning with a unified text-to-text transformer},
  author  = {Raffel, Colin and Shazeer, Noam and Roberts, Adam and Lee, Katherine and Narang, Sharan and Matena, Michael and Zhou, Yanqi and Li, Wei and Liu, Peter J},
  journal = {Journal of machine learning research},
  volume  = {21},
  number  = {140},
  pages   = {1--67},
  year    = {2020}
}

@article{xu2021instance,
  title   = {Instance-dependent partial label learning},
  author  = {Xu, Ning and Qiao, Congyu and Geng, Xin and Zhang, Min-Ling},
  journal = {Advances in Neural Information Processing Systems},
  volume  = {34},
  pages   = {27119--27130},
  year    = {2021}
}

@inproceedings{ghosh2021contrastive,
  title     = {Contrastive learning improves model robustness under label noise},
  author    = {Ghosh, Aritra and Lan, Andrew},
  booktitle = {Proceedings of the IEEE/CVF Conference on Computer Vision and Pattern Recognition},
  pages     = {2703--2708},
  year      = {2021}
}

@inproceedings{li2021detecting,
  title     = {Detecting the fake candidate instances: Ambiguous label learning with generative adversarial networks},
  author    = {Li, Changchun and Li, Ximing and Ouyang, Jihong and Wang, Yiming},
  booktitle = {Proceedings of the 30th ACM International Conference on Information \& Knowledge Management},
  pages     = {903--912},
  year      = {2021}
}

@inproceedings{zhang2021exploiting,
  title     = {Exploiting class activation value for partial-label learning},
  author    = {Zhang, Fei and Feng, Lei and Han, Bo and Liu, Tongliang and Niu, Gang and Qin, Tao and Sugiyama, Masashi},
  booktitle = {International conference on learning representations},
  year      = {2021}
}

@inproceedings{wang2022pico,
  title     = {Pico: Contrastive label disambiguation for partial label learning},
  author    = {Wang, Haobo and Xiao, Ruixuan and Li, Yixuan and Feng, Lei and Niu, Gang and Chen, Gang and Zhao, Junbo},
  booktitle = {International conference on learning representations},
  year      = {2022}
}

@inproceedings{wu2022revisiting,
  title        = {Revisiting consistency regularization for deep partial label learning},
  author       = {Wu, Dong-Dong and Wang, Deng-Bao and Zhang, Min-Ling},
  booktitle    = {International conference on machine learning},
  pages        = {24212--24225},
  year         = {2022},
  organization = {PMLR}
}

@article{ruan2022biased,
  title     = {Biased Complementary-Label Learning Without True Labels},
  author    = {Ruan, Jianfei and Zheng, Qinghua and Zhao, Rui and Dong, Bo},
  journal   = {IEEE Transactions on Neural Networks and Learning Systems},
  volume    = {35},
  number    = {2},
  pages     = {2616--2627},
  year      = {2022},
  publisher = {IEEE}
}

@inproceedings{li2022selective,
  title     = {Selective-supervised contrastive learning with noisy labels},
  author    = {Li, Shikun and Xia, Xiaobo and Ge, Shiming and Liu, Tongliang},
  booktitle = {Proceedings of the IEEE/CVF conference on computer vision and pattern recognition},
  pages     = {316--325},
  year      = {2022}
}

@inproceedings{xu2022webly,
  title     = {Webly-Supervised Fine-Grained Recognition with Partial Label Learning.},
  author    = {Xu, Yu-Yan and Shen, Yang and Wei, Xiu-Shen and Yang, Jian},
  booktitle = {IJCAI},
  pages     = {1502--1508},
  year      = {2022}
}

@inproceedings{lyu2022deep,
  title     = {Deep Graph Matching for Partial Label Learning.},
  author    = {Lyu, Gengyu and Wu, Yanan and Feng, Songhe},
  booktitle = {IJCAI},
  pages     = {3306--3312},
  year      = {2022}
}

@inproceedings{xia2022ambiguity,
  title     = {Ambiguity-Induced Contrastive Learning for Instance-Dependent Partial Label Learning.},
  author    = {Xia, Shiyu and Lv, Jiaqi and Xu, Ning and Geng, Xin},
  booktitle = {IJCAI},
  pages     = {3615--3621},
  year      = {2022}
}

@article{qiao2022decompositional,
  title   = {Decompositional generation process for instance-dependent partial label learning},
  author  = {Qiao, Congyu and Xu, Ning and Geng, Xin},
  journal = {arXiv preprint arXiv:2204.03845},
  year    = {2022}
}

@inproceedings{kim2022diffusionclip,
  title     = {Diffusionclip: Text-guided diffusion models for robust image manipulation},
  author    = {Kim, Gwanghyun and Kwon, Taesung and Ye, Jong Chul},
  booktitle = {Proceedings of the IEEE/CVF conference on computer vision and pattern recognition},
  pages     = {2426--2435},
  year      = {2022}
}

@inproceedings{preechakul2022diffusion,
  title     = {Diffusion autoencoders: Toward a meaningful and decodable representation},
  author    = {Preechakul, Konpat and Chatthee, Nattanat and Wizadwongsa, Suttisak and Suwajanakorn, Supasorn},
  booktitle = {Proceedings of the IEEE/CVF conference on computer vision and pattern recognition},
  pages     = {10619--10629},
  year      = {2022}
}

@article{zhao2022egsde,
  title   = {Egsde: Unpaired image-to-image translation via energy-guided stochastic differential equations},
  author  = {Zhao, Min and Bao, Fan and Li, Chongxuan and Zhu, Jun},
  journal = {Advances in Neural Information Processing Systems},
  volume  = {35},
  pages   = {3609--3623},
  year    = {2022}
}

@article{hertz2022prompt,
  title   = {Prompt-to-prompt image editing with cross attention control},
  author  = {Hertz, Amir and Mokady, Ron and Tenenbaum, Jay and Aberman, Kfir and Pritch, Yael and Cohen-Or, Daniel},
  journal = {arXiv preprint arXiv:2208.01626},
  year    = {2022}
}

@article{kwon2022diffusion,
  title   = {Diffusion models already have a semantic latent space},
  author  = {Kwon, Mingi and Jeong, Jaeseok and Uh, Youngjung},
  journal = {arXiv preprint arXiv:2210.10960},
  year    = {2022}
}

@inproceedings{xu2023progressive,
  title        = {Progressive purification for instance-dependent partial label learning},
  author       = {Xu, Ning and Liu, Biao and Lv, Jiaqi and Qiao, Congyu and Geng, Xin},
  booktitle    = {International Conference on Machine Learning},
  pages        = {38551--38565},
  year         = {2023},
  organization = {PMLR}
}

@inproceedings{he2023candidate,
  title     = {Candidate-aware selective disambiguation based on normalized entropy for instance-dependent partial-label learning},
  author    = {He, Shuo and Yang, Guowu and Feng, Lei},
  booktitle = {Proceedings of the IEEE/CVF International Conference on Computer Vision},
  pages     = {1792--1801},
  year      = {2023}
}

@inproceedings{brooks2023instructpix2pix,
  title     = {Instructpix2pix: Learning to follow image editing instructions},
  author    = {Brooks, Tim and Holynski, Aleksander and Efros, Alexei A},
  booktitle = {Proceedings of the IEEE/CVF Conference on Computer Vision and Pattern Recognition},
  pages     = {18392--18402},
  year      = {2023}
}

@inproceedings{pan2023effective,
  title     = {Effective real image editing with accelerated iterative diffusion inversion},
  author    = {Pan, Zhihong and Gherardi, Riccardo and Xie, Xiufeng and Huang, Stephen},
  booktitle = {Proceedings of the IEEE/CVF International Conference on Computer Vision},
  pages     = {15912--15921},
  year      = {2023}
}

@inproceedings{tumanyan2023plug,
  title     = {Plug-and-play diffusion features for text-driven image-to-image translation},
  author    = {Tumanyan, Narek and Geyer, Michal and Bagon, Shai and Dekel, Tali},
  booktitle = {Proceedings of the IEEE/CVF Conference on Computer Vision and Pattern Recognition},
  pages     = {1921--1930},
  year      = {2023}
}

@inproceedings{wang2023stylediffusion,
  title     = {Stylediffusion: Controllable disentangled style transfer via diffusion models},
  author    = {Wang, Zhizhong and Zhao, Lei and Xing, Wei},
  booktitle = {Proceedings of the IEEE/CVF International Conference on Computer Vision},
  pages     = {7677--7689},
  year      = {2023}
}

@inproceedings{wu2023uncovering,
  title     = {Uncovering the disentanglement capability in text-to-image diffusion models},
  author    = {Wu, Qiucheng and Liu, Yujian and Zhao, Handong and Kale, Ajinkya and Bui, Trung and Yu, Tong and Lin, Zhe and Zhang, Yang and Chang, Shiyu},
  booktitle = {Proceedings of the IEEE/CVF conference on computer vision and pattern recognition},
  pages     = {1900--1910},
  year      = {2023}
}

@inproceedings{wallace2023edict,
  title     = {Edict: Exact diffusion inversion via coupled transformations},
  author    = {Wallace, Bram and Gokul, Akash and Naik, Nikhil},
  booktitle = {Proceedings of the IEEE/CVF Conference on Computer Vision and Pattern Recognition},
  pages     = {22532--22541},
  year      = {2023}
}

@article{epstein2023diffusion,
  title   = {Diffusion self-guidance for controllable image generation},
  author  = {Epstein, Dave and Jabri, Allan and Poole, Ben and Efros, Alexei and Holynski, Aleksander},
  journal = {Advances in Neural Information Processing Systems},
  volume  = {36},
  pages   = {16222--16239},
  year    = {2023}
}

@article{li2023layerdiffusion,
  title   = {LayerDiffusion: Layered Controlled Image Editing with Diffusion Models},
  author  = {Li, Pengzhi and Huang, QInxuan and Ding, Yikang and Li, Zhiheng},
  journal = {arXiv e-prints},
  pages   = {arXiv--2305},
  year    = {2023}
}

@article{li2023moecontroller,
  title   = {MoEController: Instruction-based Arbitrary Image Manipulation with Mixture-of-Expert Controllers},
  author  = {Li, Sijia and Chen, Chen and Lu, Haonan},
  journal = {arXiv preprint arXiv:2309.04372},
  year    = {2023}
}

@inproceedings{patashnik2023localizing,
  title     = {Localizing object-level shape variations with text-to-image diffusion models},
  author    = {Patashnik, Or and Garibi, Daniel and Azuri, Idan and Averbuch-Elor, Hadar and Cohen-Or, Daniel},
  booktitle = {Proceedings of the IEEE/CVF International Conference on Computer Vision},
  pages     = {23051--23061},
  year      = {2023}
}

@article{ho2023diffusion,
  title   = {Diffusion-ss3d: Diffusion model for semi-supervised 3d object detection},
  author  = {Ho, Cheng-Ju and Tai, Chen-Hsuan and Lin, Yen-Yu and Yang, Ming-Hsuan and Tsai, Yi-Hsuan},
  journal = {Advances in Neural Information Processing Systems},
  volume  = {36},
  pages   = {49100--49112},
  year    = {2023}
}

@article{you2024diffusion,
  title   = {Diffusion models and semi-supervised learners benefit mutually with few labels},
  author  = {You, Zebin and Zhong, Yong and Bao, Fan and Sun, Jiacheng and Li, Chongxuan and Zhu, Jun},
  journal = {Advances in Neural Information Processing Systems},
  volume  = {36},
  year    = {2024}
}

@inproceedings{tian2024crosel,
  title     = {CroSel: Cross Selection of Confident Pseudo Labels for Partial-Label Learning},
  author    = {Tian, Shiyu and Wei, Hongxin and Wang, Yiqun and Feng, Lei},
  booktitle = {Proceedings of the IEEE/CVF Conference on Computer Vision and Pattern Recognition},
  pages     = {19479--19488},
  year      = {2024}
}

@inproceedings{wu2024distilling,
  title     = {Distilling Reliable Knowledge for Instance-Dependent Partial Label Learning},
  author    = {Wu, Dong-Dong and Wang, Deng-Bao and Zhang, Min-Ling},
  booktitle = {Proceedings of the AAAI Conference on Artificial Intelligence},
  volume    = {38},
  pages     = {15888--15896},
  year      = {2024}
}

@article{huang2024diffusion,
  title   = {Diffusion model-based image editing: A survey},
  author  = {Huang, Yi and Huang, Jiancheng and Liu, Yifan and Yan, Mingfu and Lv, Jiaxi and Liu, Jianzhuang and Xiong, Wei and Zhang, He and Chen, Shifeng and Cao, Liangliang},
  journal = {arXiv preprint arXiv:2402.17525},
  year    = {2024}
}

@article{xu2024cyclenet,
  title   = {Cyclenet: Rethinking cycle consistency in text-guided diffusion for image manipulation},
  author  = {Xu, Sihan and Ma, Ziqiao and Huang, Yidong and Lee, Honglak and Chai, Joyce},
  journal = {Advances in Neural Information Processing Systems},
  volume  = {36},
  year    = {2024}
}

@inproceedings{guo2024focus,
  title     = {Focus on your instruction: Fine-grained and multi-instruction image editing by attention modulation},
  author    = {Guo, Qin and Lin, Tianwei},
  booktitle = {Proceedings of the IEEE/CVF Conference on Computer Vision and Pattern Recognition},
  pages     = {6986--6996},
  year      = {2024}
}

@inproceedings{geng2024instructdiffusion,
  title     = {Instructdiffusion: A generalist modeling interface for vision tasks},
  author    = {Geng, Zigang and Yang, Binxin and Hang, Tiankai and Li, Chen and Gu, Shuyang and Zhang, Ting and Bao, Jianmin and Zhang, Zheng and Li, Houqiang and Hu, Han and others},
  booktitle = {Proceedings of the IEEE/CVF Conference on Computer Vision and Pattern Recognition},
  pages     = {12709--12720},
  year      = {2024}
}

@inproceedings{zhang2024hive,
  title     = {Hive: Harnessing human feedback for instructional visual editing},
  author    = {Zhang, Shu and Yang, Xinyi and Feng, Yihao and Qin, Can and Chen, Chia-Chih and Yu, Ning and Chen, Zeyuan and Wang, Huan and Savarese, Silvio and Ermon, Stefano and others},
  booktitle = {Proceedings of the IEEE/CVF Conference on Computer Vision and Pattern Recognition},
  pages     = {9026--9036},
  year      = {2024}
}

@inproceedings{huberman2024edit,
  title     = {An edit friendly ddpm noise space: Inversion and manipulations},
  author    = {Huberman-Spiegelglas, Inbar and Kulikov, Vladimir and Michaeli, Tomer},
  booktitle = {Proceedings of the IEEE/CVF Conference on Computer Vision and Pattern Recognition},
  pages     = {12469--12478},
  year      = {2024}
}

@inproceedings{brack2024ledits++,
  title     = {Ledits++: Limitless image editing using text-to-image models},
  author    = {Brack, Manuel and Friedrich, Felix and Kornmeier, Katharia and Tsaban, Linoy and Schramowski, Patrick and Kersting, Kristian and Passos, Apolin{\'a}rio},
  booktitle = {Proceedings of the IEEE/CVF Conference on Computer Vision and Pattern Recognition},
  pages     = {8861--8870},
  year      = {2024}
}

@inproceedings{bodur2024iedit,
  title     = {iedit: Localised text-guided image editing with weak supervision},
  author    = {Bodur, Rumeysa and Gundogdu, Erhan and Bhattarai, Binod and Kim, Tae-Kyun and Donoser, Michael and Bazzani, Loris},
  booktitle = {Proceedings of the IEEE/CVF Conference on Computer Vision and Pattern Recognition},
  pages     = {7426--7435},
  year      = {2024}
}

@article{zhang2024magicbrush,
  title   = {Magicbrush: A manually annotated dataset for instruction-guided image editing},
  author  = {Zhang, Kai and Mo, Lingbo and Chen, Wenhu and Sun, Huan and Su, Yu},
  journal = {Advances in Neural Information Processing Systems},
  volume  = {36},
  year    = {2024}
}

@article{chen2024label,
  title   = {Label-retrieval-augmented diffusion models for learning from noisy labels},
  author  = {Chen, Jian and Zhang, Ruiyi and Yu, Tong and Sharma, Rohan and Xu, Zhiqiang and Sun, Tong and Chen, Changyou},
  journal = {Advances in Neural Information Processing Systems},
  volume  = {36},
  year    = {2024}
}

@article{xu2024variational,
  title     = {Variational Label Enhancement for Instance-Dependent Partial Label Learning},
  author    = {Xu, Ning and Qiao, Congyu and Zhao, Yuchen and Geng, Xin and Zhang, Min-Ling},
  journal   = {IEEE Transactions on Pattern Analysis and Machine Intelligence},
  year      = {2024},
  publisher = {IEEE}
}

@article{li2024learning,
  title   = {Learning Causal Transition Matrix for Instance-dependent Label Noise},
  author  = {Li, Jiahui and Chang, Tai-Wei and Kuang, Kun and Li, Ximing and Chen, Long and Zhou, Jun},
  journal = {arXiv preprint arXiv:2412.13516},
  year    = {2024}
}

@article{wang2025realistic,
  title   = {Realistic evaluation of deep partial-label learning algorithms},
  author  = {Wang, Wei and Wu, Dong-Dong and Wang, Jindong and Niu, Gang and Zhang, Min-Ling and Sugiyama, Masashi},
  journal = {arXiv preprint arXiv:2502.10184},
  year    = {2025}
}

@inproceedings{yang2025mixed,
  title     = {Mixed blessing: Class-wise embedding guided instance-dependent partial label learning},
  author    = {Yang, Fuchao and Cheng, Jianhong and Liu, Hui and Dong, Yongqiang and Jia, Yuheng and Hou, Junhui},
  booktitle = {Proceedings of the 31st ACM SIGKDD Conference on Knowledge Discovery and Data Mining V. 1},
  pages     = {1763--1772},
  year      = {2025}
}
